# Exploring Key Factors for Long-Term Vessel Incident Risk Prediction

**Tianyi Chen**

Department of Civil and Environmental Engineering, National Univesity of Singapore,
Singapore 117576, Singapore
tychen@nus.edu.sg

**Hua Wang**

School of Automotive and Transportation Engineering, Hefei University of Technology,
Hefei 230009, China
hwang191901@hfut.edu.cn

**Yutong Cai**

Department of Civil and Environmental Engineering, National Univesity of Singapore,
Singapore 117576, Singapore
ceecaiy@nus.edu.sg

**Maohan Liang**

Department of Civil and Environmental Engineering, National Univesity of Singapore,
Singapore 117576, Singapore
mhliang@nus.edu.sg

**Qiang Meng**

Department of Civil and Environmental Engineering, National Univesity of Singapore,
Singapore 117576, Singapore
ceemq@nus.edu.sg
(Corresponding Author)

## ABSTRACT

Factor analysis acts a pivotal role in enhancing maritime safety. Most previous studies conduct factor analysis within the framework of incident-related label prediction, where the developed models can be categorized into short-term and long-term prediction models. The long-term models offer a more strategic approach, enabling more proactive risk management, compared to the short-term ones. Nevertheless, few studies have devoted to rigorously identifying the key factors for the long-term prediction and undertaking comprehensive factor analysis. Hence, this study aims to delve into the key factors for predicting the incident risk levels in the subsequent year given a specific datestamp. The majority of candidate factors potentially contributing to the incident risk are collected from vessels' historical safety performance data spanning up to five years. An improved embedded feature selection method, which integrates Random Forest classifier with a feature filtering process, is proposed to identify key risk-contributing factors from the candidate pool. A dataset with information of 131,652 vessels collected from 2015 to 2023 is utilized for case study. The results demonstrate superior performances of the proposed method in incident prediction and factor interpretability. Comprehensive analysis is conducted upon the key factors, which could help maritime stakeholders formulate management strategies for incident prevention.

## 1. INTRODUCTION

Vessel incidents, posing risks of substantial loss of life and property damage, have consistently been a focal point of concern in the maritime industry. The European Maritime Safety Agency (EMSA) reported that there were in total 23,814 casualties and incidents from 2014 to 2022, which resulted in 184 ships lost and 604 fatalities [1]. Therefore, it is imperative to explore and comprehend the incident risk-contributing factors, which can offer insights into incident causation and guidance for incident prevention. In the literature, a couple of studies have endeavored to investigate the factors potentially contributing maritime incidents. Most of these studies conducted factor analysis based on short-term incident prediction models. These short-term prediction models typically use the factors collected when an incident is occurring or about to occur, such as the factors related to human behaviors, environment, and navigation, to predict or estimate incident-related labels, such as incident type, consequence, or severity. The short-term prediction models are able to reveal how the factors contribute to an individual incident from a microscopic perspective, which helps create a safer operating environment for vessels. Nevertheless, the short-term prediction has two-fold limitations: first, the time window between prediction and incident occurrence is short and uncertain, potentially insufficient for risk alleviation measures; and second, some of the factors may not be easily collected in real-time.

As compared to the short-term prediction models, the long-term prediction models extends the timeframe for incident prediction. Typically, the long-term prediction models predict incident-related labels in the near future primarily based on the vessel's historical safety-related records (e.g., incidents, deficiencies, etc.). The likelihood of a vessel being involved in an incident in the coming future (e.g., in the subsequent year) is usually associated with its historical safety performances. For example, in 2024, a cargo ship collided with a major bridge in Baltimore, the United States, resulting in the collapse of the bridge. It was found that this ship had a serious collision with a wall at the Port of Antwerp in Belgium in 2016, the ship owner has faced four lawsuits due to worker injuries since 2018, and this ship was cited for propulsion deficiency in Chile in 2023 [2]. This case demonstrates that the ship's histrocal risk records may reflect the drawbacks in its operation and management, potentially increasing the likelihood of future incidents. The long-term prediction is conducted from a macroscopic perspective and circumvents the limitations of the short-term prediction, which provides maritime stakeholders and governmental agencies with the insights into the vessel's overall safety level, enabling the development of effective safety enhancement strategies accordingly. Therefore, this study aims to conduct factor analysis based on the long-term maritime incident prediction.

Possibly due to the difficulty of accessing data, limited studies have focused on the long-term maritime incident prediction, with many lacking rigor in factor identification. Specifically, in these studies, there are two major gaps that remain to be resolved: first, most studies overlooked a fact that the contributions of the historical factors to the future incident risk may decay over time; and second, in most studies, the factors are represented by a constant format, which may underrate the informativeness of the factors. Besides, the causation of a maritime incident can be complicated, which may involve numerous potential factors from multiple aspects including vessel profiles, regulatory compliance, operational practices, etc. Hence, to bridge the gaps, more potential factors should be taken into consideration. However, most of these potential factors may be noisy, redundant, and less informative, which can adversely affect prediction performance and factor interpretability. Therefore, it is essential to identify the key factors that significantly contribute to the incident risk from the potential factors. As an effective feature selection method, the embedded method with Random Forest classifier [3] as the host machine learning model has been widely utilized to identify key factors and conduct factor analysis in both academia and industry. This method selects the key factors based on the feature importance of each potential factor, which is calculated as per information-theoretic criterion during the training process of Random Forest classifier. Nevertheless, the method does not take into consideration the correlations between factors, which may reduce prediction performance and explainability of the selected factors, especially when handling a large number of potential factors [4].

This study considers the factors related to vessel's historical records and profile information. The factors from the historical records are categorized into incidents, Port State Control (PSC) deficiencies, detentions, sailing, Document of Compliance (DOC) performances, and flag state performances. The incident risk label is defined by grading risk levels based on the number of incidents and incident severity in the subsequent year. To resolve the aforementioned gaps and challenges, this study introduces temporal decay factors and adopts various time windows for collecting potential historical factors, and these historical factors are represented using various formatting approaches (e.g., annual basis and cumulative basis, with and without decay factor, etc.). Besides, this study improved the conventional Random Forest-based feature selection method by incorporating a feature filtering process to mitigate the impact of higly-correlated factors and enhance rationality and interpretability of the selected key factors. The analysis of the key factors can provide a more comprehensive understanding of the underlying causes and vulnerabilities of vessel incidents in the long term. By understanding the key factors, port authorities, shipping companies, and other stakeholders could implement incident preventive measures, strengthen safety protocols, and mitigate risks before they escalate into incidents.

The remainder of this paper is organized as follows. Section 2 reviews the related literature and summarizes the research gaps to fill in our work. Section 3 introduces the sources of data and the construction of the labelled dataset utilized for feature selection. Section 4 explains the proposed feature selection method. Section 5 presents a case study. Section 6 concludes the paper.

## 2. LITERATURE REVIEW

### 2.1 Factor analysis on maritime incidents

Factor analysis indeed plays a significant role in maritime incident investigation by assisting in the identification and understanding of the underlying factors contributing to incidents. This study has reviewed a total of 31 papers that investigated maritime incidents. These papers were published in reputable journals in the domain of transportation and maritime in the past decade. These papers are summarized in Table A1 in Appendix A, which includes factor categories, label definition, analytical methodology, and scope of each study. Herein, the factor categories indicate the types of the factors (e.g., vessel particular, human factor, etc.) that may contribute to a maritime incident, the label definitions denote the types of incident-related label (e.g., risk levels, incident types, etc.), the analytical methodologies refer to the models used to elucidate the relation between factors and label, and the scopes are the research boundaries or constraints of the reviewed paper. Most of these papers aimed to predict or assess the incident-related label based on the selected factors, thereby exploring the contributions of these factors to the label. The Sankey diagram as shown in Figure 1 visually summarizes these previous studies in terms of factor categories, label identification, and analytical methodologies.

The models proposed in these previous studies are categorized into short-term prediction and long-term prediction models based on the time window during which the factors are collected. The short-term prediction models refer to the models that utilize the factors collected exactly when incidents occur, such as environmental factors, human factors, navigational factors, etc.

Most of these models were built using data from historical incident reports, and the factors are collected at the time of the incident as per the reports. However, when these models are applied in reality, the factors may be observed in advance before the incident occurrence. For example, some environmental factors (e.g., weather, sea condition, etc.) and navigational factors (e.g., vessel speed, vessel traffic, etc.) may remain consistent when an incident is occurring and when it is about to occur. Therefore, these models can be utilized to conduct short-term prediction, but the extent to which they can predict in advance may vary and not be certain. Furthermore, of note, there is a special sub-category within the short-term prediction models, referred to as incident assessment model. While some incident assessment models are named as 'prediction model' as well in some studies, those models indeed incorporate incident particulars (e.g., date & time, location, number of injuries/deaths, etc.) as factors, which make them not suitable for prediction purposes.

As illustrated in Figure 1, environmental factors, vessel particulars, human factors, and incident particulars are the top four categories of factors that have been frequently investigated in these reviewed studies. This manifests that most of these studies used short-term prediction models. As listed under the 'label definitions' in Figure 1, incident types and incident severity level are the two dominant types of label adopted for prediction, followed by incident risk level, incident consequences, and risk probability. The short-term prediction models can be used to investigate the factors that have significant contributions to incidents and explore causality of an incident. Specifically, with incident types as label, the prediction models can reveal both commonalities and heterogeneities among incidents. Besides, with incident severity indicators (e.g., severity level, risk level, consequences, etc.) as label, the prediction models are able to identify specific risk-contributing factors that may increase the likelihood of severe incidents and uncover the underlying causality that may exacerbate the consequence of incidents. Although those models can help raise awareness of various risk-contributing factors upon the prediction, they have the following two limitations when utilized in reality: 1) the short and uncertain window between identifying incident risks and potential incident occurrence may pose a challenge for making timely interventions and proactive risk management; and 2) some of the factors, especially the human factors (e.g., psychological status, fatigue, distraction, etc.) and management-related factors (e.g., decision support, emergency plan, leadership violation, etc.) may not be precisely collected in real-time.

As compared with the short-term prediction models, the long-term prediction models provide incident prediction and assessment from a general and macroscopic perspective. Typically, the long-term prediction models are used to estimate the likelihood of a vessel being involved in a incident in the near future based upon the factors including its previous histories (e.g., incidents, detention, deficiencies, etc.) and the historical performances of its management entities (e.g., shipping companies, flag states, etc.). The long-term prediction models are more practical than the short-term prediction models in reality, as they do not have the aforementioned limitations associated with the short-term prediction. Further, the factor analysis based upon the long-term prediction can provide governmental agencies with insights into the overall safety level of a vessel, which can also serve as a criterion for commercial entities in selecting shipping vessel. As compared to the short-term prediction, only a few previous studies [5-8] focused on the long-term prediction, and the summary of these previous studies can be found in Table A1. These studies attempted to investigate the contributions of vessel's historical safety-related factors to the likelihood of incident occurrences, employing various definitions of incident label. However, these studies are to some extent less rigorous, with the following research gaps that remain to be addressed.

1) Most studies roughly predefined a fixed time period for collecting all the historical factors, but the durations of impact may vary across those factors. For example, an incident and a detention that occurred one year ago may exert different effects on the current safety status of a vessel. Similarly, the impact of an incident that occurred one year ago may differ from that of an incident that occurred two years ago on the current safety status. However, few studies have taken into consideration such heterogeneity in the impact duration within the historical factors.
2) Most studies merely used the count of a particular type of events within the predefined time period as a factor. For example, the count of inspection deficiencies in the past three years is used as factor in a previous study. However, there are several ways to represent vessel's historical performances on inspection deficiencies, such as the average deficiencies over the three years, the annual deficiency count incorporating time weight, etc. However, few studies have attempted to explore the most representative formatting approach for a factor.

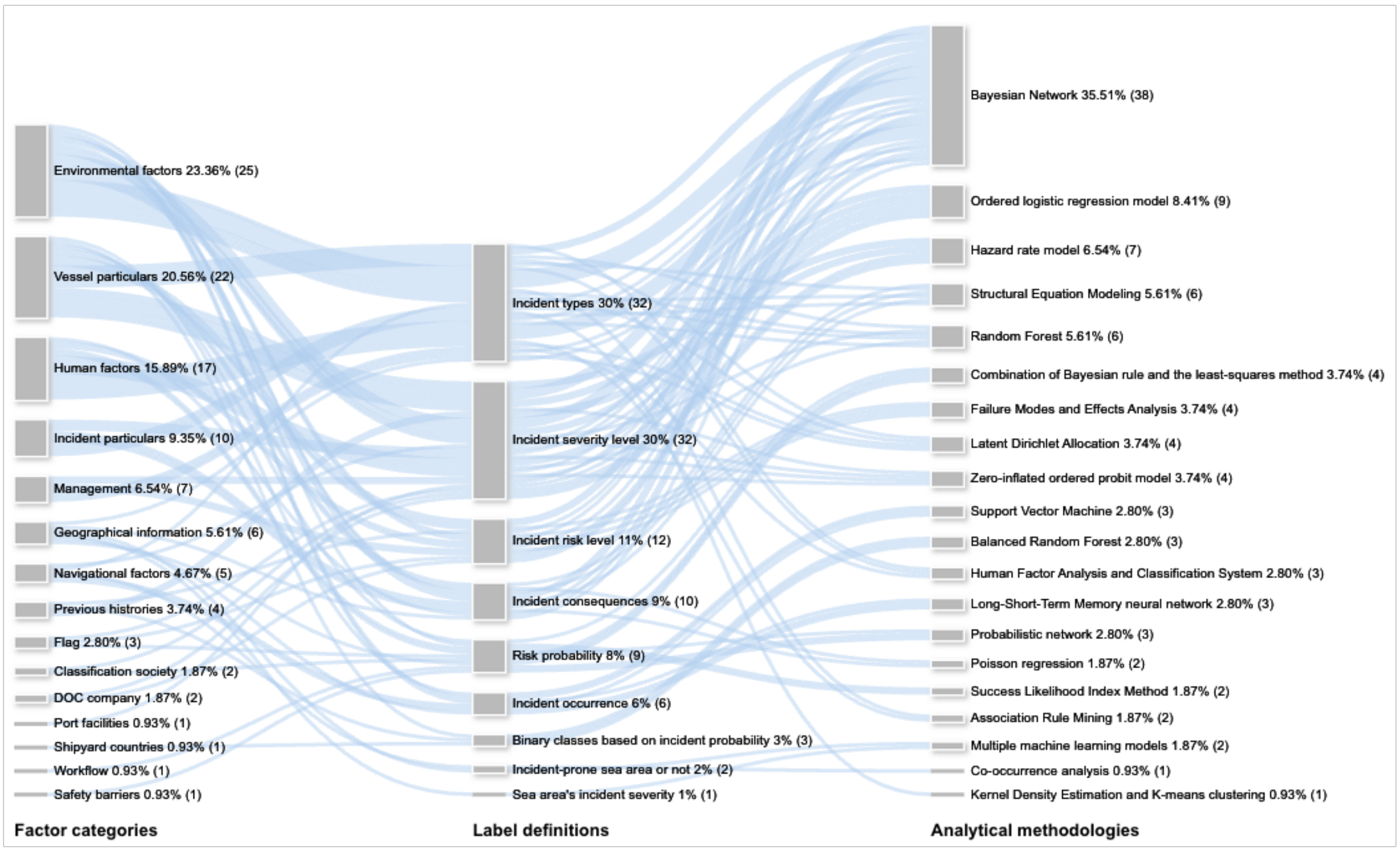

Figure 1 Sankey diagram of previous maritime incident studies

## 2.2 Feature selection methods

From the list of analytical methodologies in Figure 1, it is evident that in recent years, machine learning methods (e.g., Bayesian Network, Random Forest, Support Vector Machine, etc.) have gained prominence in investigating maritime incident, leveraging their capabilities of handling complex data and enhancing predictive performance. The machine learning methods have been used to predict incident-related label, measure factor importance, and explain incident causality in previous studies. However, most studies adopted a great number of factors when building a machine learning model and failed to consider the correlations between those factors. Some of those factors might be irrelevant and redundant, which may cause noise and reduce prediction performance of machine learning model. Also, the factors with high correlation between each other may introduce redundant information and increase model complexity, which may further lead to overfitting problem and reduce the interpretability of model prediction. To this end, it is of necessity to implement feature selection (i.e., factor selection) to identify the most relevant and informative factors while discarding irrelevant, redundant, and noisy ones when building machine learning models. This is particularly crucial when dealing with numerous factors that could potentially contribute to maritime incidents.

Given a labelled dataset where each sample has a series of factors and a label, there are three main categories of feature selection methods, namely, filter, wrapper, and embedded methods, which are commonly used for identifying key factors [9]. Herein, the filter methods select factors based on their relevance to and the label, as measured by correlation coefficient, mutual information, chi-square test, etc. The wrapper methods assess factor subsets by training and testing machine learning models on different subsets of factors, using the performances of the models as the criteria. The embedded methods incorporate feature selection into the model training process, where the importance of each factor is determined during the training process based on the built-in mechanisms or regularization of the model. The embedded method indeed is a combination of the filter and wrapper methods, which also integrates the advantages of both methods. The embedded methods are able to eliminate the need for separating the factors into subsets, capture complex factor interaction and dependencies effectively, and prioritize the most informative factors while minimizing the impact of the irrelevant ones. These capabilities make the embedded methods more efficient and accurate while being less prone to over-fitting in diverse applications across different domains [10].

The method with Random Forest classifier as the host machine learning model is one of the most popular and widely used embedded methods. Random Forest classifier is computationally efficient and scalable, making it suitable for coping with large datasets with high-dimensional feature space. Besides, Random Forest classifier is robust to the variance and bias in the data, as it aggregates predictions from multiple decision trees and such an aggregation reduces the risk of individual tree memorizing noise in the training data. Hence, this method can effectively identify stable and representative factors, thereby demonstrating satisfactory feature selection performances [11-13]. In the domain of maritime studies, Random Forest classifier has been widely used for feature selection and factor analysis in recent years. For example, Soner et al. [14] used the classifier to measure the importance of the factors potentially contributing to ship performance under operating conditions. Zhang et al. [4] used the classifier to capture the features that represent the similarity between vessel trajectories. Knapp and van de Velden [8] applied the classifier to measure the relevance of factors to ship-specific incident probabilities. Li et al. [15] adopted the classifier to predict maritime incident consequences and identify the factors that significantly contribute to the consequence. Kandel and Baroud [16] used the classifier to predict type of Arctic incidents and identify their risk factors. Most of these studies employed the feature importance generated during the training process of the classifier to evaluate the contributions of factors to the label. However, they hardly consider the correlations among the factors, which may deteriorate the performances in feature selection especially when handling the dataset with a large amount of correlated or similar factors [4].

## 3. DATA PREPARATION

### 3.1 Data landscape

The data used for this study are provided by RightShip, which is an Environmental, Social, and Governance (ESG) company dedicated to setting global benchmarks for maritime industry. The data include basic profiles, operational status, and event records of all the vessels that are accessible by RightShip from 2015 to 2023. Specifically, the data are elaborated as below:

- **Incident records**: The recorded incidents include various incident types, such as collision, fire, explosion, etc. The data not only provide the essential information of each incident (e.g., time, location, number of fatalities and injuries, etc.), but also classifies the incidents into three categories, namely, Categories A, B, and C, based on severity. Herein, Category A incidents refer to the incidents that cause a loss of life, total loss (i.e., vessel is completely destroyed or lost), or very serious consequences. Category B incidents are those that result in significant damage to the vessel or ship unseaworthiness. Category C incidents are those with no significant damage to the vessel and the vessel still remains seaworthy.
- **PSC detention and deficiency records:** Detention is an outcome of a PSC inspection when the inspector determines that the identified deficiencies are serious enough to detain the vessel at port for a certain period of time. Deficiencies are the items identified by inspectors during PSC inspections as an immediate threat to the vessel, its personnel, or the environment. The data include the information such as time, duration, location (i.e., port), etc. of each detention and number of deficiencies in each PSC inspection.
- **Vessel's basic information:** The data include various information related to the vessel itself, such as the vessel's locations collected by Automatic Identification System (AIS), the details of the DOC company that manages and operates the vessel, the demerits of flag state identified by the International Chamber of Shipping (ICS), the physical profile of the vessel, etc. The alterations of the vessel's information, such as the changes of DOC company or flag state, are also recorded in the data.

### 3.2 Factor extraction and label definition

For feature selection, it is necessary to have a labeled dataset with each sample having multiple factors and a target label. In this study, regarding a vessel at a specified datestamp as a sample, the factors are extracted based on information related to the vessel's historical performances in the past five years from the given datestamp. The aggregated severity of all incidents involving the vessel in the next year is defined as the label. To obtain the key factors, it is essential to build a pool of candidate factors that may potentially contribute to the label. The candidate factors are extracted from the data introduced in Section 3.1. Figure 2 shows the categorization of the candidate factors, where yellow, green, and blue blocks denote primary, secondary, and tertiary categories of factors, respectively. Each tertiary category is denoted by a notation (e.g., C1) and followed an example of the candidate factors that belong to the category.

The candidate factors are grouped into seven categories, namely, incidents, PSC deficiencies, detentions, sailing, DOC performances, flag performances, and profile information. Excluding the profile information, the candidate factors from the remaining categories are extracted over the past five years from the given datestamp. In most cases, the factors from those categories of historical performances are presented in annual, cumulative, or decayed cumulative formats. The three formats can be referred to through the examples following the tertiary categories. It is worth noting that the factor in the decayed cumulative format is calculated considering decay factors. This is because that, for some factors, the same performances in earlier and more recent periods may have different effects on the current situation. For example, given that the decay factor of the past $i^{th}$ year is $k_i$ and the annual total numbers of PSC deficiencies of a vessel in the past $i^{th}$ year is $M_i$, the decayed cumulative number of deficiencies in the past $n$ years from the given datestamp equals $\sum_i^n k_i \cdot M_i$. According to the empirical values used by RighShip, the decay factors for the past five years (i.e., $\{k_1, k_2, k_3, k_4, k_5\}$) are set as $\{5, 4, 3, 3, 2\}$ in this study. The primary categories are introduced as follows:

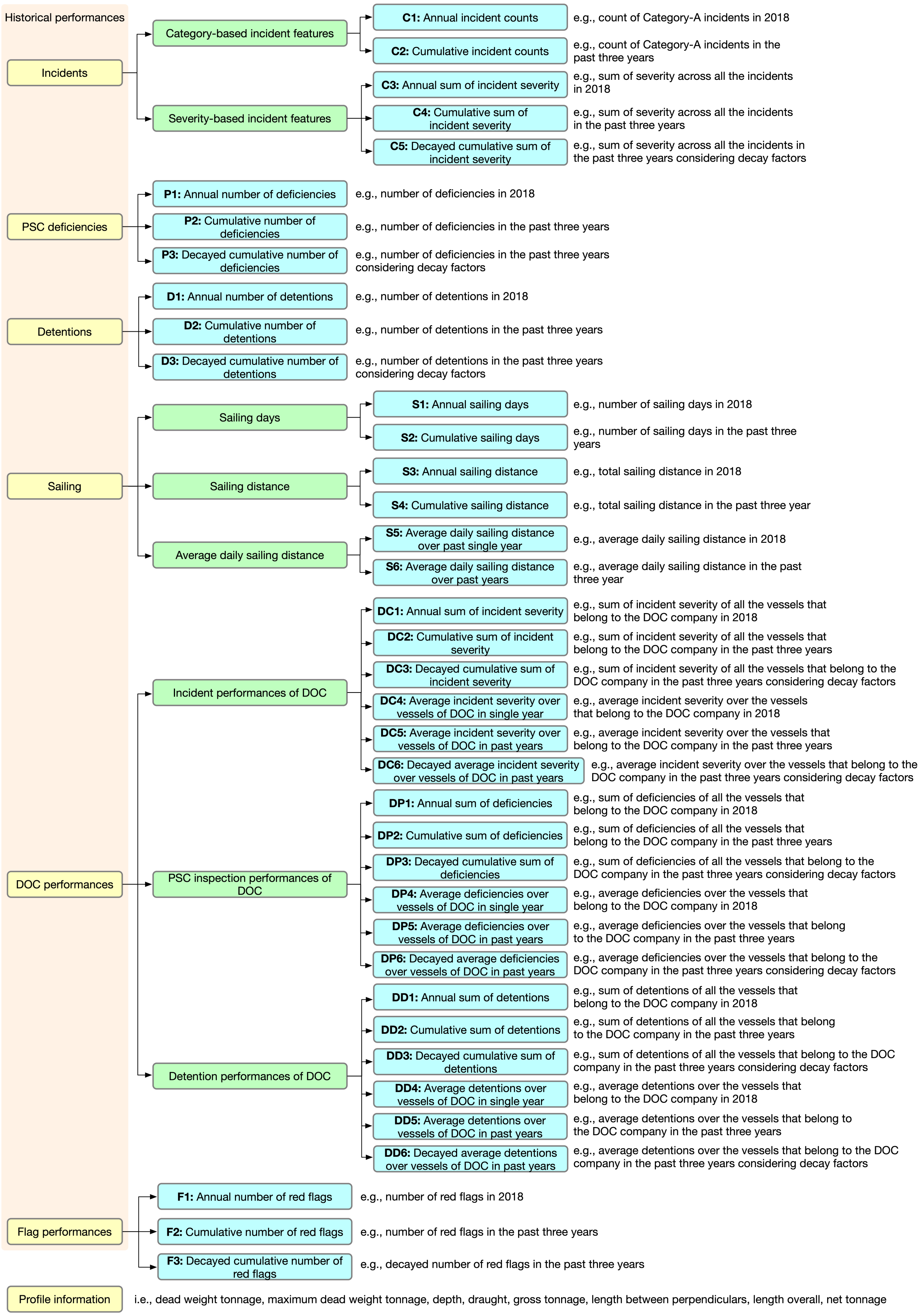

Figure 2 Categorization of candidate factors

- **Incidents:** The factors from this category can be further divided into category-based and severity-based incident factors. The category-based incident factors are extracted according to aforementioned severity incident categories (i.e., Categories A, B, and C). The severity-based incident factor are calculated based on quantification of incident severity. For example, the annual sum of incident severity of a vessel is calculated as $\sum_{i}^{A,B,C} w_i\beta_i$, where $\beta_i$ is the annual count of an incident category and $w_i$ is the weight assigned to the category. As suggested by RightShip, the weights assigned to Categories A, B, and C are 6, 2, and 1 respectively in this study (i.e., $w_A = 6$, $w_B = 2$, and $w_C = 1$).
- **PSC deficiencies:** The factors are calculated based on the annual count of deficiencies identified by PSC inspection.
- **Detentions:** The factors are calculated based on the annual count of times that a vessel is detained by PSC inspection.
- **Sailing:** The factors are extracted from the AIS data of the vessel. The sailing distance is calculated based on the vessel's daily sailing distance, while the sailing days are the days when the daily sailing distance is non-zero. The average daily sailing distance over a time period equals the cumulative sailing distance divided by the sailing days.
- **DOC performances:** The factors from this categories are determined based on multiple performance indicators, including the sum of incident severity, the total number of deficiencies identified by PSC inspection, and the total number of detentions of all the vessels in the DOC company that operates the sample vessel during a given time period. Herein, the incident severity is also calculated with weights of 6, 2, and 1 respectively corresponding to Categories A, B, and C incidents. The fleet size used for calculating average incident severity, deficiencies, and intentions over the vessels in the DOC company takes into account the vessels that change DOC company. For example, given a time period, if the fleet size of a DOC company is $s_1$ for $t_1$ days and $s_2$ for another $t_2$ days, the fleet size during this period is equal to $(s_1t_1 + s_2t_2)/(t_1 + t_2)$. Similarly, if a vessel changes its DOC company during a given time period, the vessel's factors of this category are calculated by aggregating the duration weighted average performances over all the DOC companies operating the vessel during this period.

- **Flag performances:** Flag performances are calculated based on the demerits of flag state, which are reported in the flag state performance tables published by the ICS. The tables assess the performances of each flag state from multiple perspectives, including whether the flag state is on white or black lists of Paris Memorandum of Understanding (MoU), average age of fleet, attendance of International Maritime Organization (IMO) meetings, etc. If a negative performance is identified from an aspect, the flag state will receive a red flag. The factors in this category are calculated based on the annual count of the red flags that the vessel's flag state receive. Of note, similar to the factors of DOC performances, if a vessel changes its flag state during a given time period, the vessel's factors from this category are also calculated considering temporally weighted average.
- **Profile information:** The factors in this category reflect physical characteristics of the vessel's profile, which comprises dead weight tonnage (PF1), maximum dead weight tonnage (PF2), depth (PF3), draught (PF4), gross tonnage (PF5), length between perpendiculars (PF6), length overall (PF7), and net tonnage (PF8).

Since incidents are rare events, it is pretty challenging to predict whether a vessel will be involved in an incident or determine the category of incident that a vessel might be involved. In this case, employing occurrence of incident or occurrence of incident from a designated category as the label might reduce prediction accuracy and influence the results of feature selection. Instead, given a specified datestamp, this study uses the incident risk level in the forthcoming year as the label to represent the incident likelihood of a vessel. As aforementioned, the total incident severity over a year is calculated as $\sum_i^{A,B,C} w_i\beta_i$. In this way, the total incident severity of each vessel in the following year can be obtained, and then the total incident risk is classified into three levels: low, medium, and high risk levels, with interval as $\{0\}$, $(0,3)$, and $[3,+\infty)$ in terms of total incident severity, respectively. For example, if a vessel is involved in three Category A incidents and one Category B incident in the forthcoming year, the incident severity of the vessel is equal to $3 \times 1 + 1 \times 2 = 5$, and thereby the vessel is labelled with high incident risk level. The higher total incident risk suggests that the vessel is more likely to be involved in a more serious incident or minor incidents many times.

### 3.3 Data Preprocessing

In this study, the samples are collected from four datestamps, namely, 2020-Jul-01, 2021-Jan-01, 2021-Jul-01, and 2022-Jan-01, with half-year interval between two consective datestamps.

As presented in Table 1, for each datestamp, the factors are collected within the five years before the datestamp, and the label is calculated based on the incident counts in the next year after the datestamp. After removing samples with missing or abnormal data and eliminating replicated samples, a total of 131,652 samples are obtained for feature selection. The utilization of moving window method for sample collection can enhance the size and diversity of the collected samples. Hence, the machine learning model built upon these samples can become more robust and generalizable. The sample size of each label class in the dataset is presented in Table 2. It can be found that the dataset suffers from a class imbalance problem as the sample size of the low incident risk level is significantly higher than the other two risk levels. In this case, the feature selection methods are more likely to bias towards the factors that are more discriminative for the majority class and overlook the factors important for the minority class [18]. Hence, to address this problem, SMOTE-Tomek method [19], which is an effective feature selection method that have been widely used in academia and industry, is used to resample the dataset. The resampling results are listed in Table 2 as well.

Table 1 Summary of sample collection with respect to datestamps[1]

| Datestamps | Time windows for factor collection | Time windows for label identification | Sample sizes (vessels) |
|---|---|---|---|
| 2020-Jul-01 | 2015-Jul-01to 2020-Jun-01 | 2020-Jul-01 to 2021-Jul-01 | 25,046 |
| 2021-Jan-01 | 2016-Jan-01to 2021-Jan-01 | 2021-Jan-01 to 2022-Jul-01 | 34,510 |
| 2021-Jul-01 | 2016-Jul-01to 2021-Jul-01 | 2021-Jul-01 to 2022-Jul-01 | 34,714 |
| 2022-Jan-01 | 2017-Jan-01to 2022-Jan-01 | 2022-Jan-01 to 2023-Jan-01 | 37,382 |
| Total | | | 131,652 |

1. The format of time window is specified as ‘day A to day B’, where the day A is included while the day B is excluded.

Table 2 Breakdown of sample sizes in datasets

| Datasets | Label (incident risk levels) | | | Total |
|---|---|---|---|---|
| | Low | Medium | High | |
| Original dataset | 125,892 | 5,226 | 534 | 131,652 |
| Resampled dataset | 9,500 | 5,104 | 1,445 | 16,049 |

# 4. FEATURE SELECTION METHOD

## 4.1 Method framework

As mentioned in Section 2.2, although many studies have employed the embedded method with Random Forest classifier as the host machine learning for feature selection and factor analysis, most of them failed to consider the correlations between the factors, which may influence the performances of model prediction and interpretability of selected factors. Thus, to alleviate this problem, this study proposes an improved embedded feature selection method that integrates a feature filtering process into the embedded feature selection. Figure 3 shows the framework of the proposed feature selection method. The proposed method consists of three steps: classifier training, feature filtering, and cross validation. Given the input of the resampled dataset, an initial factor rank with a descending order of feature importance is generated during the training process of the classifier (i.e., Random Forest classifier). Then, the feature filtering process is conducted to reduce the factors that are highly correlated with the others according to the initial factor rank and form a filtered factor rank. Finally, the key factors are identified in an iterative process. In each iteration, the classifier cross-validated on the resampled data with the first *n* factors in the filtered factor rank, and the prediction performances resulting from the cross-validation are recorded. The first *n* factors on which the classifier achieves the most satisfactory performance are selected as key factors. As shown in Figure 3, the red arrow line indicates the procedure of the conventional embedded method after the classifier training, which bypasses the process of feature filtering. The technical details for the three steps will be elaborated in Sections 4.2, 4.3, and 4.4, respectively.

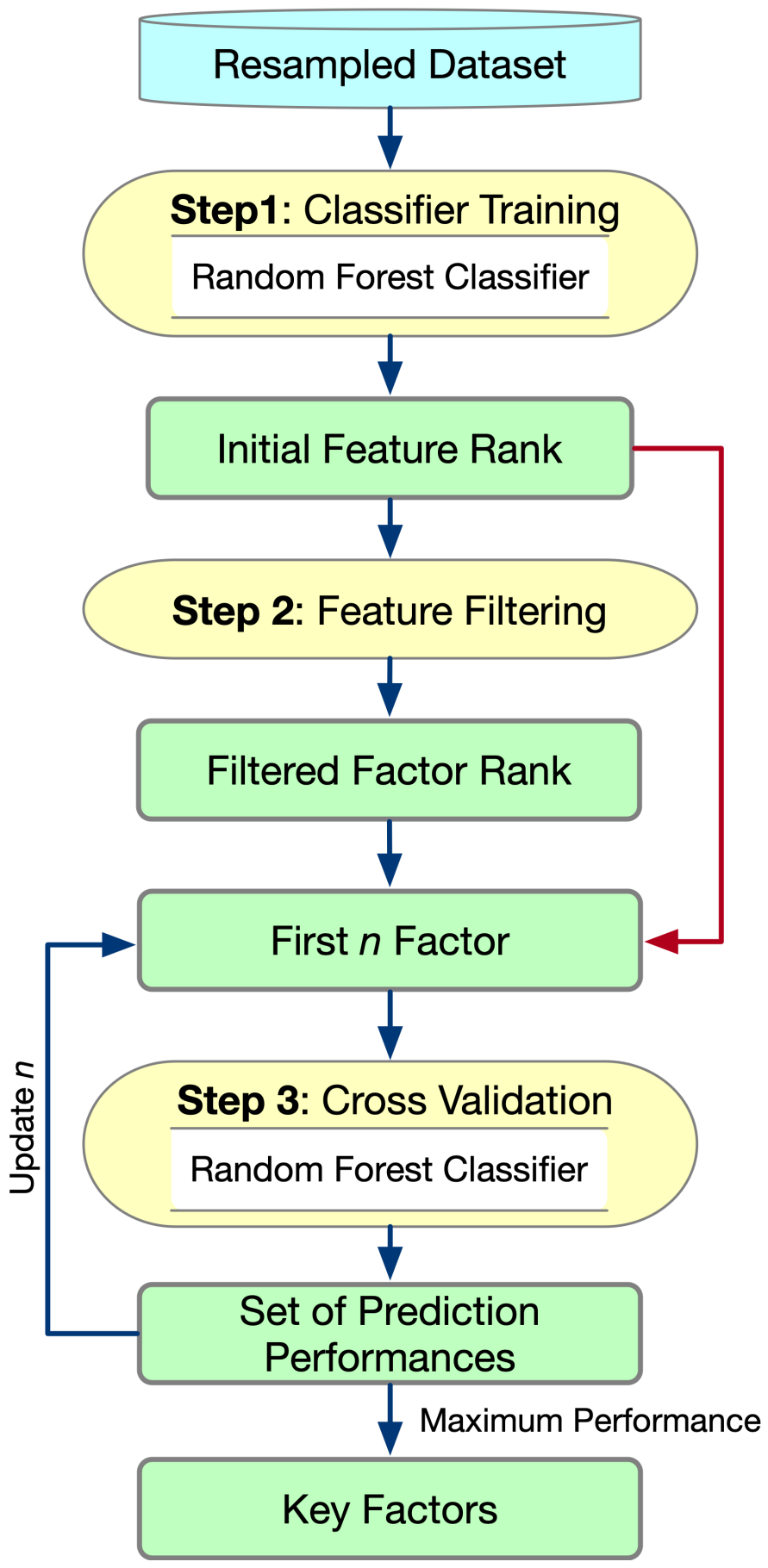

Figure 3 Framework of proposed feature selection method

### 4.2 Feature importance calculation

In the proposed feature selection method, the initial factor rank is formed based on the feature importance generated during the training process of Random Forest classifier. Random Forest classifier is an ensemble learning method that combines multiple decision trees, and the diverse set of trees ensure the robustness of the classifier, leading to more stable and reliable estimates of feature importance. The feature importance can comprehensively reflect the contribution of a factor to the label, which serves as a valuable indicator that provides insights into underlying patterns in the data. Gini impurity and information gain [3] are the two commonly used metrics for measuring feature importance in previous studies. However, these two metrics have limited quantitative interpretability, as they only provide a global measure of feature importance based on the decrease in impurity attained by splitting nodes in the decision trees. Therefore, in this study, Shapley Additive exPlanation (SHAP) value [20] is employed as the metric to measure the feature importance, which offers a remedy to the limitation of the above two conventional metrics.

As compared to the conventional metrics, SHAP values can provide individual-level and class-level explanations for model predictions, allowing users to understand the contribution of each factor to a specific sample or label class. Moreover, SHAP offers a global measure of feature importance based on the above two explanations, providing users with comprehensive insights into the the overall impact of each factor on the label. The initial factor rank is generated based on the global measures of feature importance provided by SHAP in this study. SHAP employs an additive feature attribution method to construct an explanation model for a Random Forest classifier, upon which the importance of each factor is calculated based on the concept of game theory. The explanation model is a linear function explanatory model, which is calculated as:

$$g(\boldsymbol{x}') = \phi_0 + \sum_{i=1}^{m} \phi_i x_i{}' \tag{1}$$

where $\boldsymbol{x}'$ is the set of simplified input factors, obtained by a mapping function $\boldsymbol{x}' = h_x(\boldsymbol{x})$ with $\boldsymbol{x} = (x_1, x_2, \ldots, x_m)$ indicating the set of all the factors, $\phi_0$ denotes the constant value, and $\phi_i$ refers to the feature importance (i.e., SHAP value) assigned to the factor $x_i$. The SHAP values extend the concept of Shapley values to provide explanations for Random Forest classification. The Shapley values quantify the average marginal contribution of each factor to the classifier's predictions across all possible subsets of factors using cooperative game theory. In the context of multi-class classification, the SHAP value of each factor is calculated separately for each class prediction. The individual SHAP value of the factor $i$ for the sample $j$ categorized in the class $c$ can be calculated as:

$$\phi_{ic}^{j} = \sum_{\boldsymbol{s} \subseteq \boldsymbol{x} \setminus \{i\}} \frac{|\boldsymbol{s}|!\,(|\boldsymbol{x}| - |\boldsymbol{s}| - 1)!}{|\boldsymbol{x}|!} \left(f_{c|\boldsymbol{s}}\left(x_{\boldsymbol{s}}^{i} \cup \{i\}\right) - f_{c|\boldsymbol{s}}\left(x_{\boldsymbol{s}}^{i}\right)\right) \tag{2}$$

where $\boldsymbol{s}$ indicates a given subset of factors, $f_{c|\boldsymbol{s}}\left(x_{\boldsymbol{s}}^{i} \cup \{i\}\right)$ refers to the conditional expectation of the class $c$ given the subset of factors $\boldsymbol{s}$ including the factor $i$ for the sample $j$, and $f_{c|\boldsymbol{s}}\left(x_{\boldsymbol{s}}^{i}\right)$ refers to the conditional expectation of the class $c$ given the subset of factors $\boldsymbol{s}$ excluding the factor $i$ for the sample $j$. Positive SHAP values indicate that the presence of the factor increases the classifier's prediction for the given class, while negative SHAP values indicate the opposite. The importance of a factor is obtained by aggregating the individual SHAP values of this factor across the samples and classes.

### 4.3 Feature filtering

Although Random Forest classifier is capable of dealing with the interactions and correlations between factors, its performances (i.e., prediction performances and explainability of factor ranking) are still reduced in the presence of many correlated factors in high-dimensional data (Darst et al., 2018). To this end, feature filtering is employed in the proposed feature selection method to eliminate the factors that are highly correlated to the others. The process of feature filtering engages two parameters, namely, correlation coefficient threshold and sliding range. Assuming that the initial factor rank is $\{f_1, f_2, \dots, f_\sigma\}$, a correlation coefficient matrix $\phi_{\sigma\times\sigma}$ based on the initial factor rank can be obtained. Each element in the matrix $\phi_{\sigma\times\sigma}$ is calculated by Pearson correlation coefficient. The Pearson correlation coefficient between the factors $f_i$ and $f_j$ can be obtained as:

$$r_{ij} = \frac{\sum(f_i(m) - \overline{f_i})(f_j(m) - \overline{f_j})}{\sqrt{\sum(f_i(m) - \overline{f_i})^2 \sum(f_j(m) - \overline{f_j})^2}} \tag{3}$$

where $f_\cdot(m)$ indicates the value of the factor $f_\cdot$ for the $m^{\text{th}}$ sample, and $\overline{f_\cdot}$ denotes the average value of the factor $f_\cdot$ over the given samples. Then, a sliding window is employed to filter the factors on the initial factor rank. For example, given the correlation coefficient threshold $r_\tau$ and the sliding range $w$ (i.e., the length of the sliding window), starting from the first factor $f_1$, the factor s in $\{f_2, f_3, \dots, f_{1+w}\}$ are removed from the initial factor rank if the correlation coefficient between the factor and $f_1$ exceeds $r_\tau$. Next, starting from the second factor $f_{2^*}$ on the updated factor rank, similarly, the factors in $\{f_{3^*}, f_{4^*}, \dots, f_{2^*+w}\}$ are removed from the updated factor rank if the correlation coefficient between the factor and $f_{2^*}$ exceeds $r_\tau$. The iteration ends when the sliding window cannot be moved forward, and the updated factor rank generated from the last round of iteration is output as the filtered factor rank.

The example in Figure 4 explains how feature filtering operates. In this example, initially there are $\sigma$ factors ranked in descending order of feature importance. The red rectangle denotes the sliding window, with the window length defined as 6 (i.e., $w = 6$) in this example. The sliding window starts to move from the top of the initial factor rank. In the first round of iteration, the correlation coefficients between the first factor (i.e., Factor #1) and its following five factors (i.e., Factors #2 to #6) in the window are retrieved. Factors #2 and #3 are removed as their correlation coefficients with Factor #1 exceed the $r_\tau$ set in this example. After that, a new factor rank is obtained as shown in Round #2, and the window is moved one rank place forward with Factor #4 as the new first factor. Filtering the factors in this way, the iteration ends when the window reaches the last factor (i.e., Factor #σ). The factor rank resulting from the last round of filtering (i.e., Round #n) is the output of the entire process.

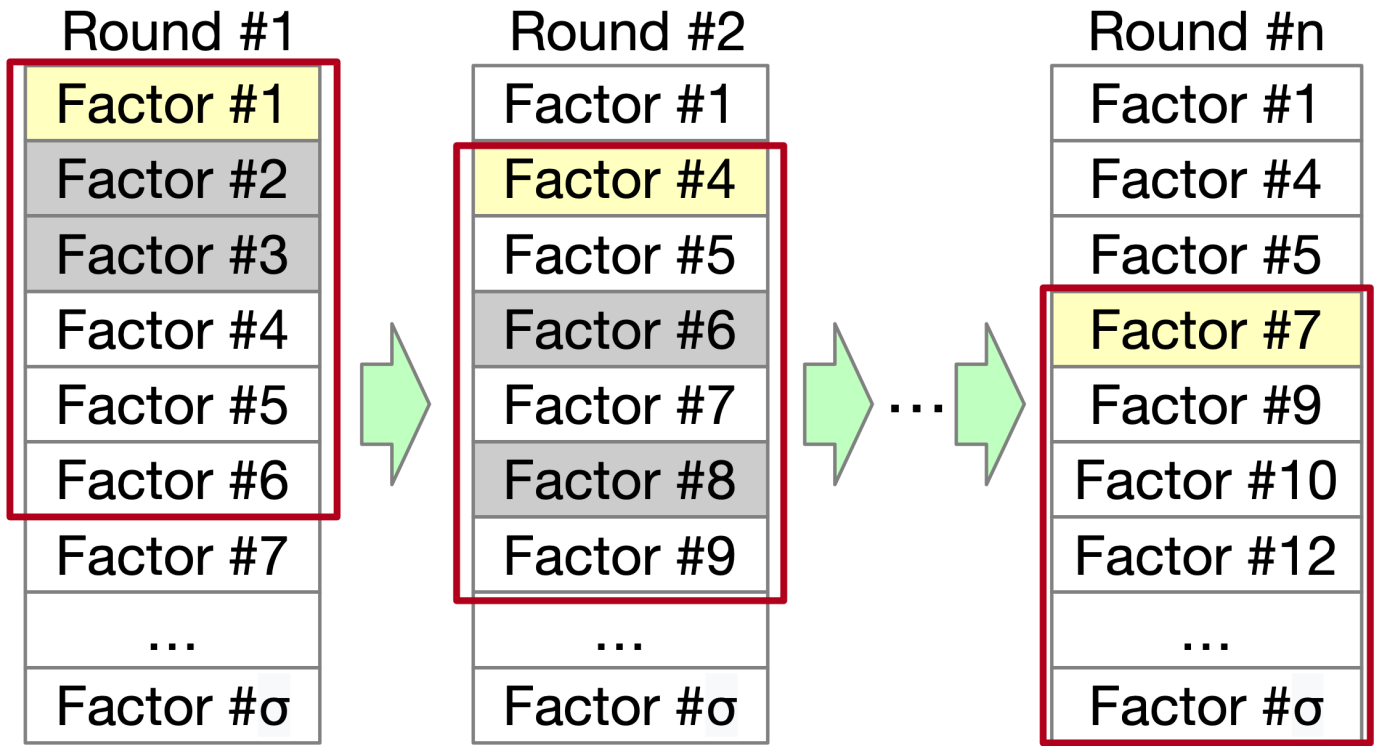

Figure 4 Example of feature filtering iteration

There are two-fold reasons for designing feature filtering in this way. First, although the high correlation between factors may significantly affect the interpretability of the selected factors, overly emphasizing the independence of the factors could result in a potential loss of critical information. The proposed feature filtering method make a trade-off between the correlations and independence of factors by adjusting the two parameters. Second, the factors that are highly correlated or similar to each other tend to be ranked closely, especially in the top-ranking area of a factor rank. Hence, considering the correlations between a factor and all other factors may not be necessary, making the incorporation of a sliding window in the method reasonable and computationally efficient. As shown in Figure 5, the two parameters are determined based upon grid searching. The cross-validation prediction performances of the Random Forest classifier trained on the resampled dataset with the filtered factors are adopted as the searching criterion. The details of performance evaluation will be introduced in Section 3.4.

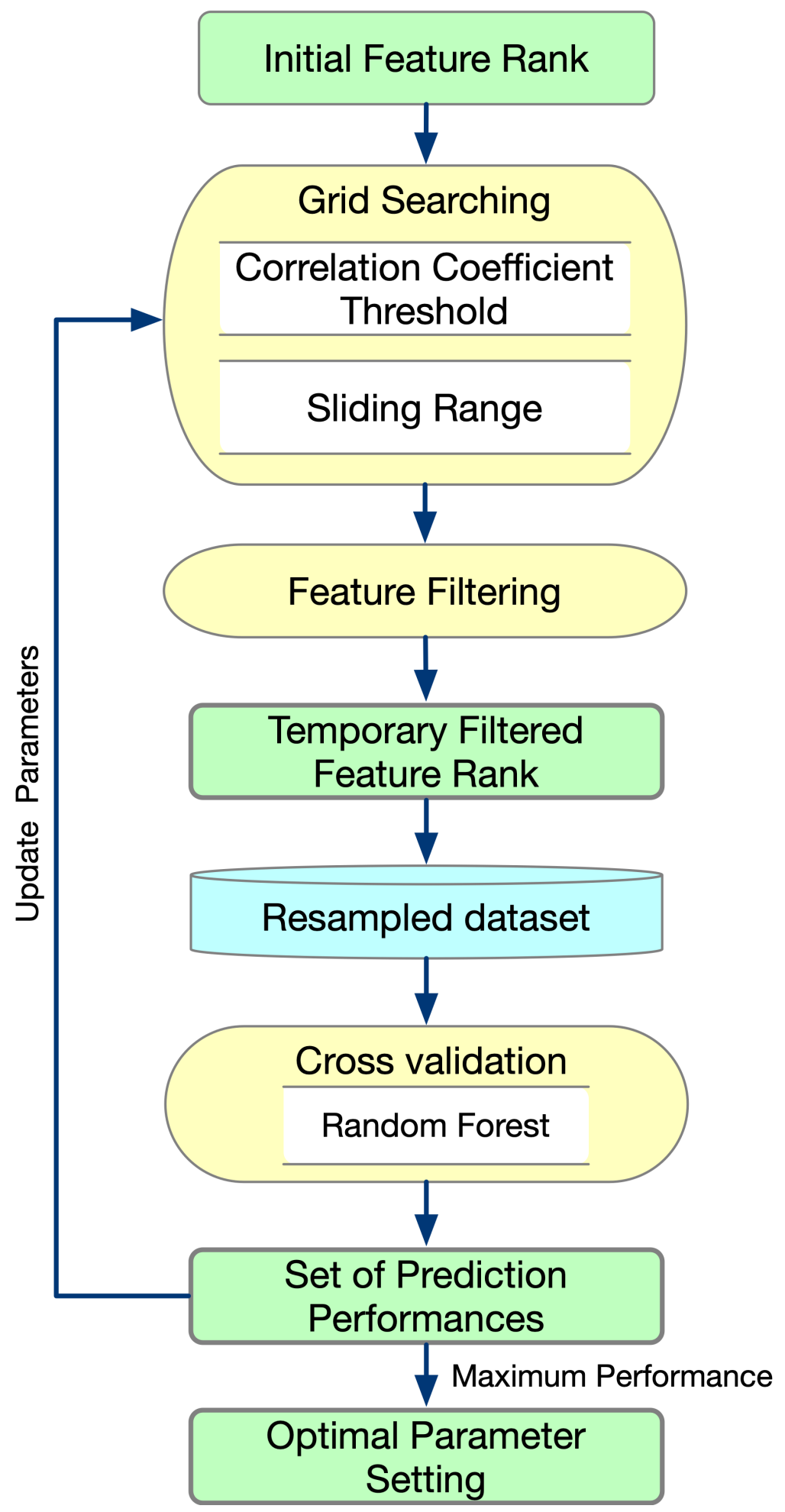

Figure 5 Scheme of searching optimal filtering parameters

### 4.4 Performance evaluation

As shown in Figures 3 and 5, the prediction performances evaluated based on cross-validation are employed as criteria to identify key factors and determine the optimal filtering parameters, respectively. The primary goal of using cross-validation is to obtain a more robust and unbiased estimate of the classifier's prediction performances, as cross-validation uses multiple subsets of the given dataset for both training and testing. As recommended by previous literature, five-fold cross-validation is employed in this study [21]. Both Figures 3 and 5 show that the performance evaluation is an iterative process. The cross-validation prediction performances are evaluated on the datasets with different designated factors, upon which the key features are identified along with the optimal parameter setting when the most satisfactory prediction performance is achieved. In a five-fold cross-validation, the classifier is trained and validated five times, with each iteration using a different fold of the dataset as the validation set while the remaining folds serve as training sets. Therefore, the prediction performances are presented as the means of the performance metrics over the five iterations. In this study, the performance metrics includes accuracy, precision, recall, f1 score, and area under the curve of receiver operating characteristics (AUC) [22]. Herein, since the resampled dataset is a multi-class dataset, the above metrics use one-against-all (OAA) scheme when measuring performance for each class, and the overall metric is calculated as the weighted average over the per-class metrics [23].

## 5. CASE STUDY

### 5.1 Feature selection

As mentioned in Section 4, to select key factors, firstly a Random Forest classifier is trained on the resampled dataset, and a factor rank with the descending order of feature importance is obtained during the training process. Then, the correlation coefficients between the candidate factors are calculated. Herein, to reduce computational expense and increase the explainability of the selected factors, this study only considers the correlation coefficient matrices for each primary category of candidate factors. Since the number of the factors from DOC performances is much larger than the other primary categories, the correlation coefficient matrices of the secondary categories under DOC performances are taken into account. Figure 6 visualizes the correlation coefficient matrix of each factor category. The notations in each sub-figure denote the tertiary categories, which can be referred to in Figure 2. It can be found that the patterns of correlation coefficient matrix are various across different factor categories. In most cases, the factors from the same tertiary category usually exhibit stronger correlation. In the categories of incidents, PSC deficiencies, and detentions, the correlation coefficients between the factors related to the cumulative performances over years are significantly higher, as compared to the factors based upon annual performances. However, in the categories under DOC performances, there are fairly high correlations among the annual performance factors.

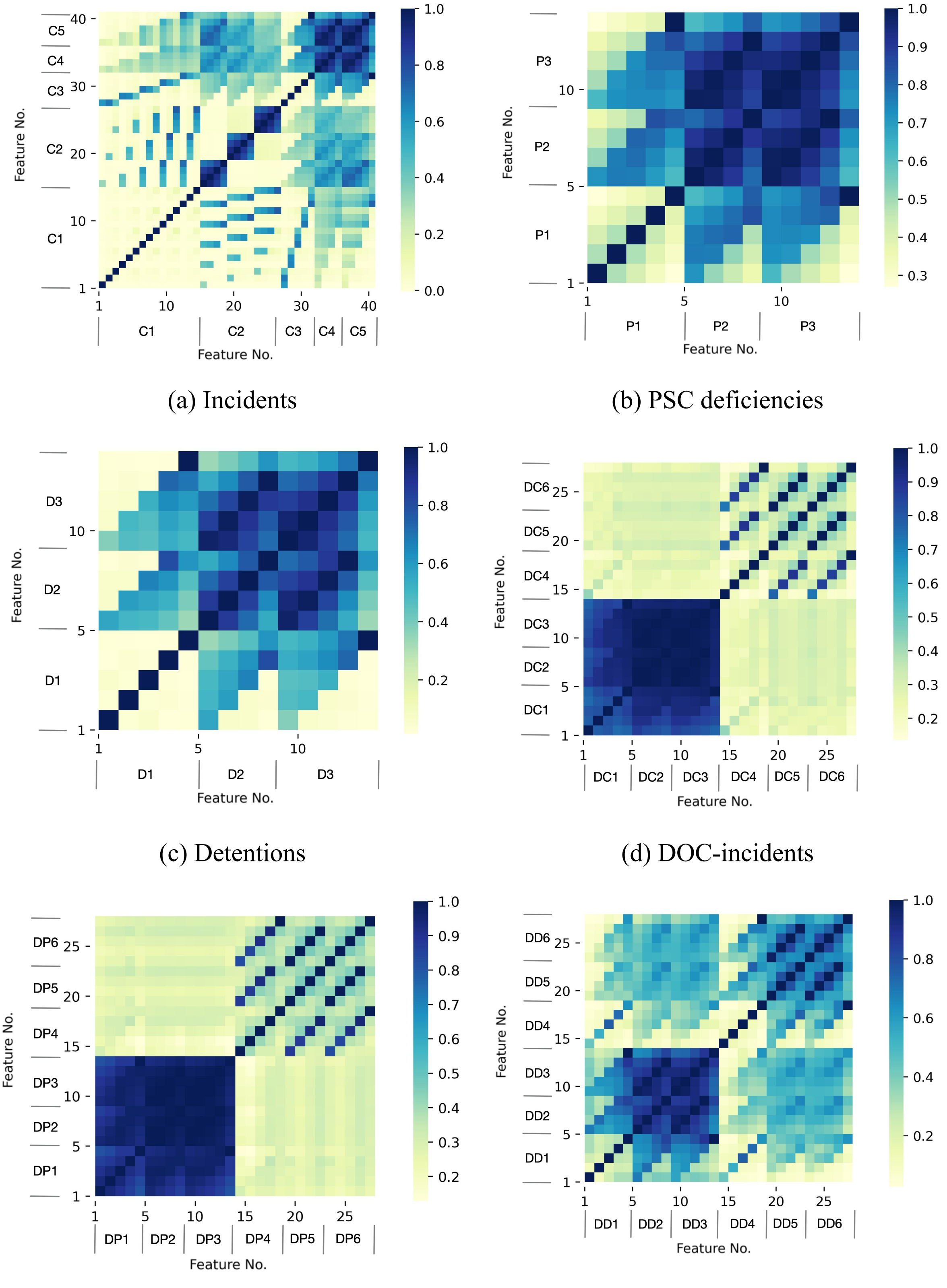

(a) Incidents (b) PSC deficiencies

(c) Detentions (d) DOC-incidents

(e) DOC-PSC inspection performances (f) DOC-detention performances

Figure 6 Correlation coefficient matrices of factor categories

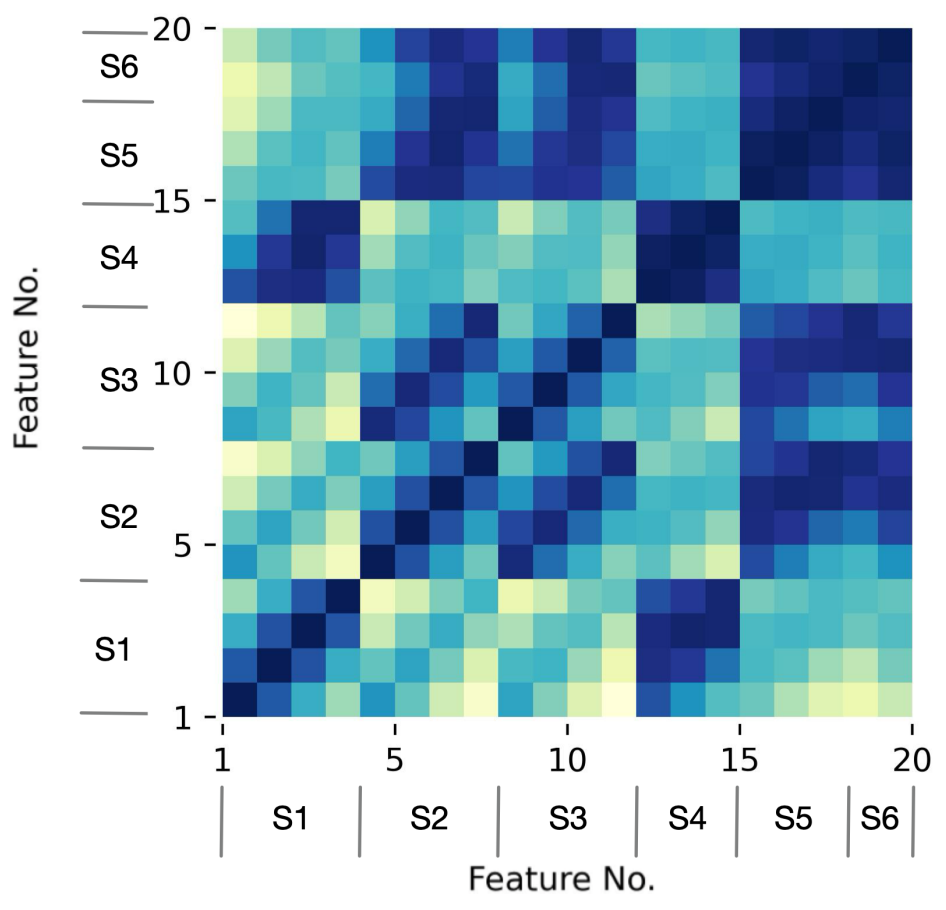

(g) Sailing

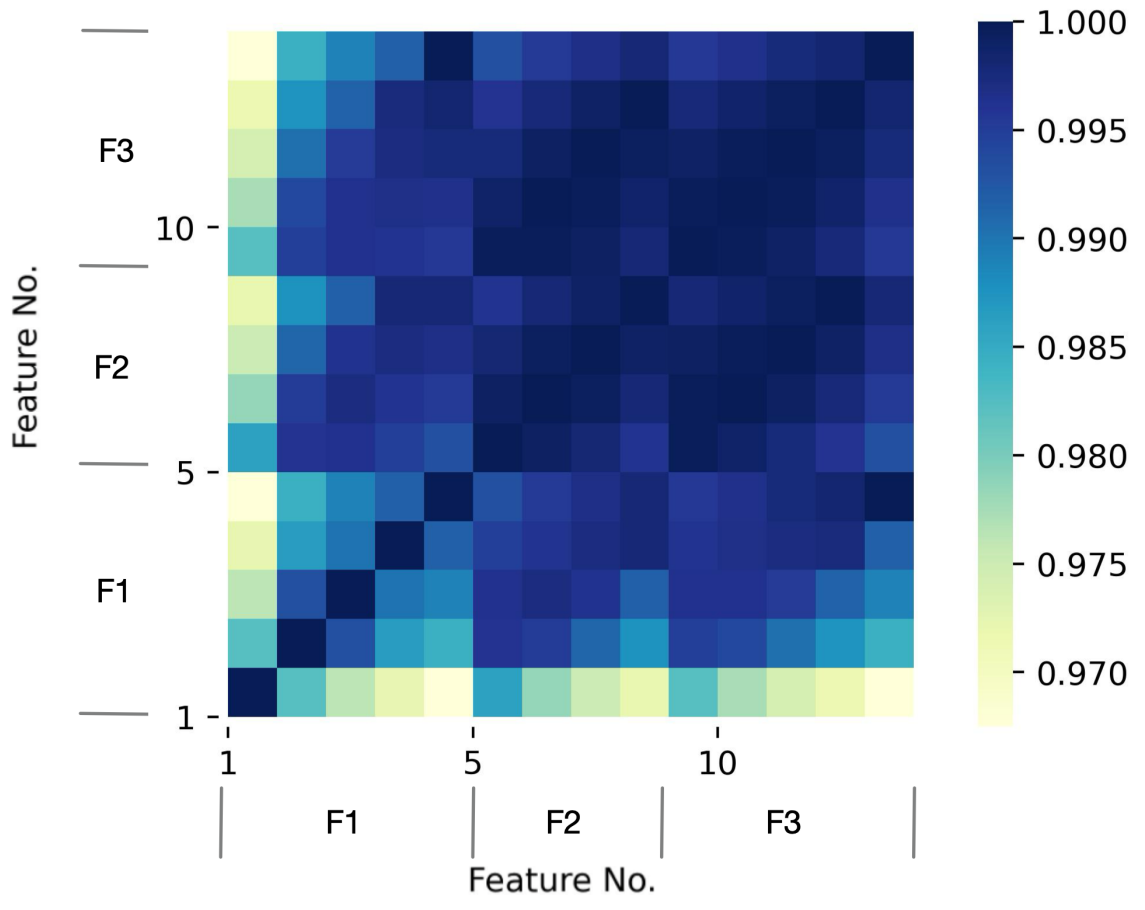

(h) Flag performances

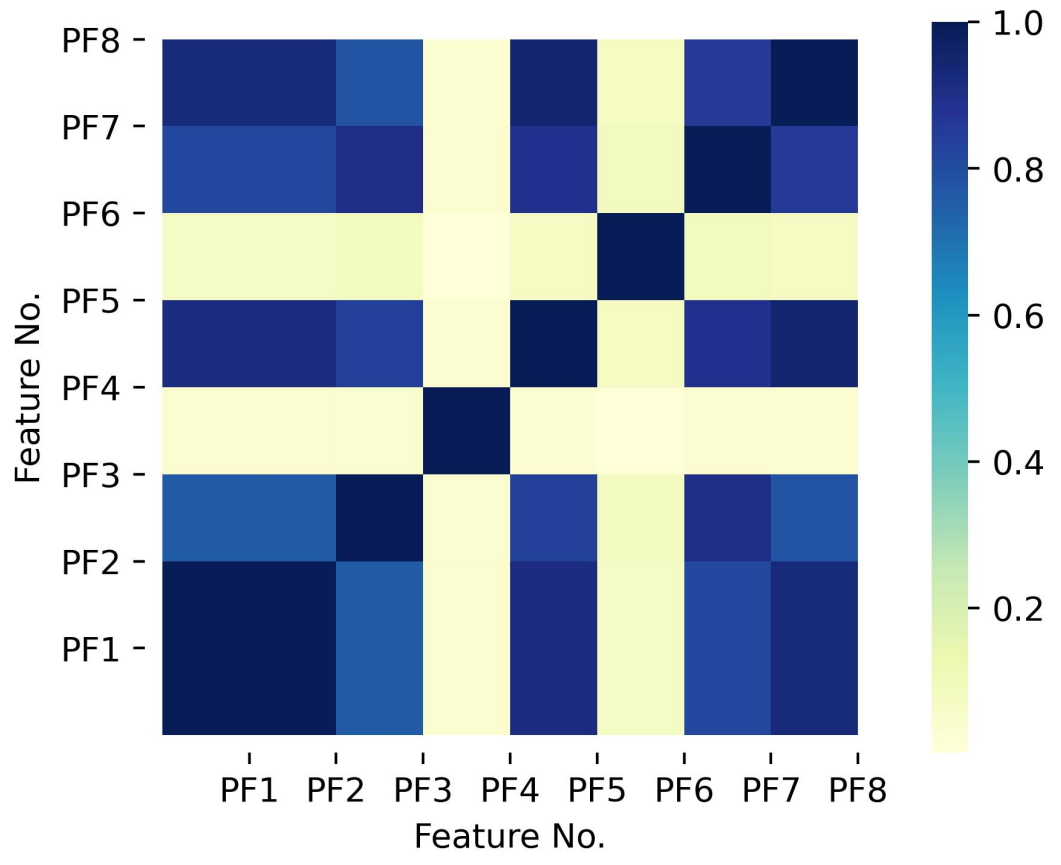

(i) Profile performances

Figure 6 Correlation coefficient matrices of factor categories (cont.)

Based upon the factor rank, the factors are further filtered using the method as introduced in Section 4.2. The two parameters for feature filtering (i.e., correlation coefficient threshold and sliding range) are identified using grid searching. Herein, the correlation coefficient threshold is searched in the range of $[0.1, 0.7]$ with 0.1 as interval, while the sliding range is searched in the range of $[5, 25]$ with 5 as interval. The five prediction performance metrics (i.e., accuracy, precision, recall, f1 score, and AUC) are employed as the criteria for identifying the parameters. Figure 7 illustrates the results of prediction performances as measured upon the grid searching. The correlation coefficient threshold is determined as 0.2, while the sliding range is determined as 15, because the maximum values in all the five metrics are achieved when employing this combined parameter setting. Besides, as shown in Figure 7, given a specified sliding range, the prediction performances are increased with the decreasing correlation coefficient threshold, and lowest prediction performances are observed when higher correlation coefficient threshold and larger sliding range are given. This manifests that high correlations between factors could negatively impact the prediction performance of machine learning classifiers, such as Random Forest classifier. Also, as shown in Figures 7(a) to 7(d), when correlation coefficient threshold is lower than 0.2 and sliding range is larger than 15, the prediction performances are kept at a consistently high level with minimal variation. As shown in Figure 7(e), given the same ranges of parameter setting, compared to the other four metrics, the prediction performance measured by AUC exhibits higher variation but is still at a high level. Plus, as presented in Figures 7(a) and 7(b), when correlation coefficient threshold is decreased from 0.2 to 0.1 and sliding range is increased from 15 to 20, it can be observed that the prediction performance is slightly reduced. This suggests that keeping the factors with certain correlations in between might contribute to higher prediction performance. Excessively increasing the independence of factors might not necessarily enhance but could potentially decrease prediction performance. This is because the excessive elimination of factors might cause the loss of crucial information, adversely affecting the predictive capabilities of classifier.

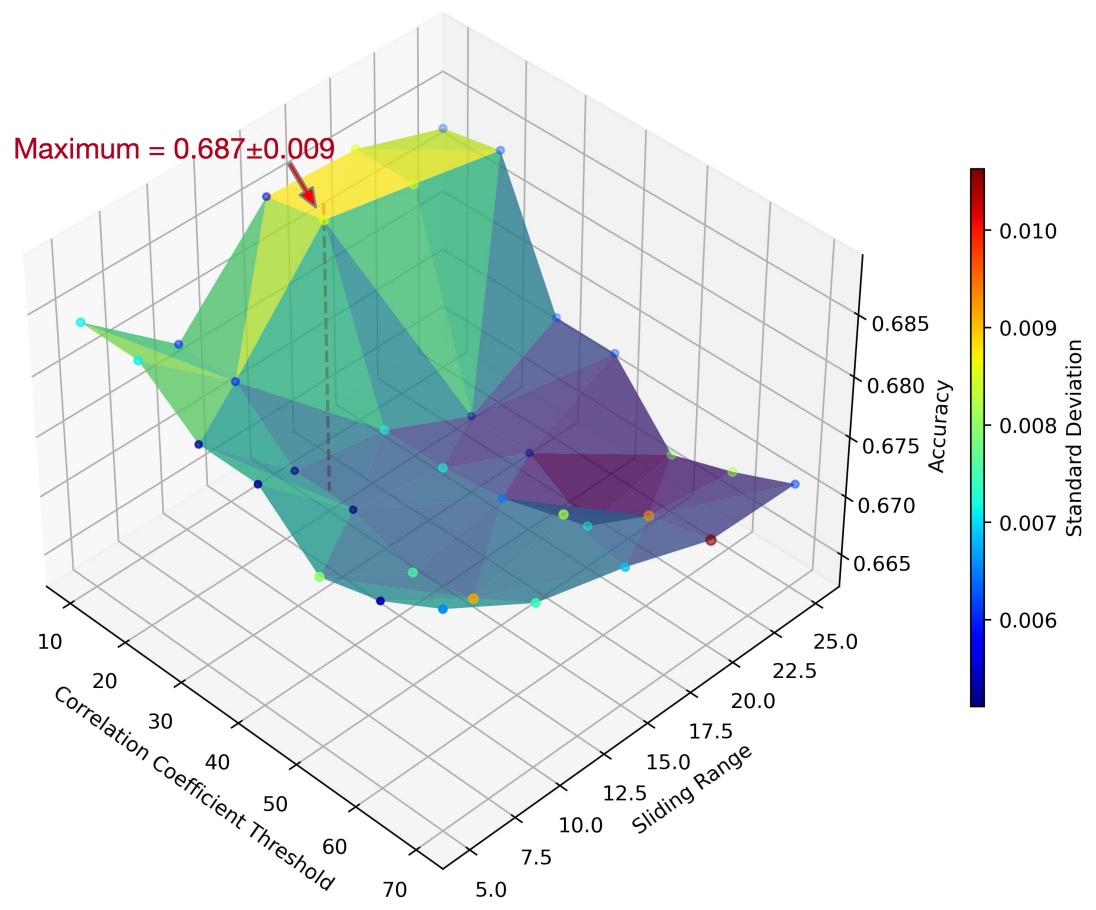

(a) Accuracy

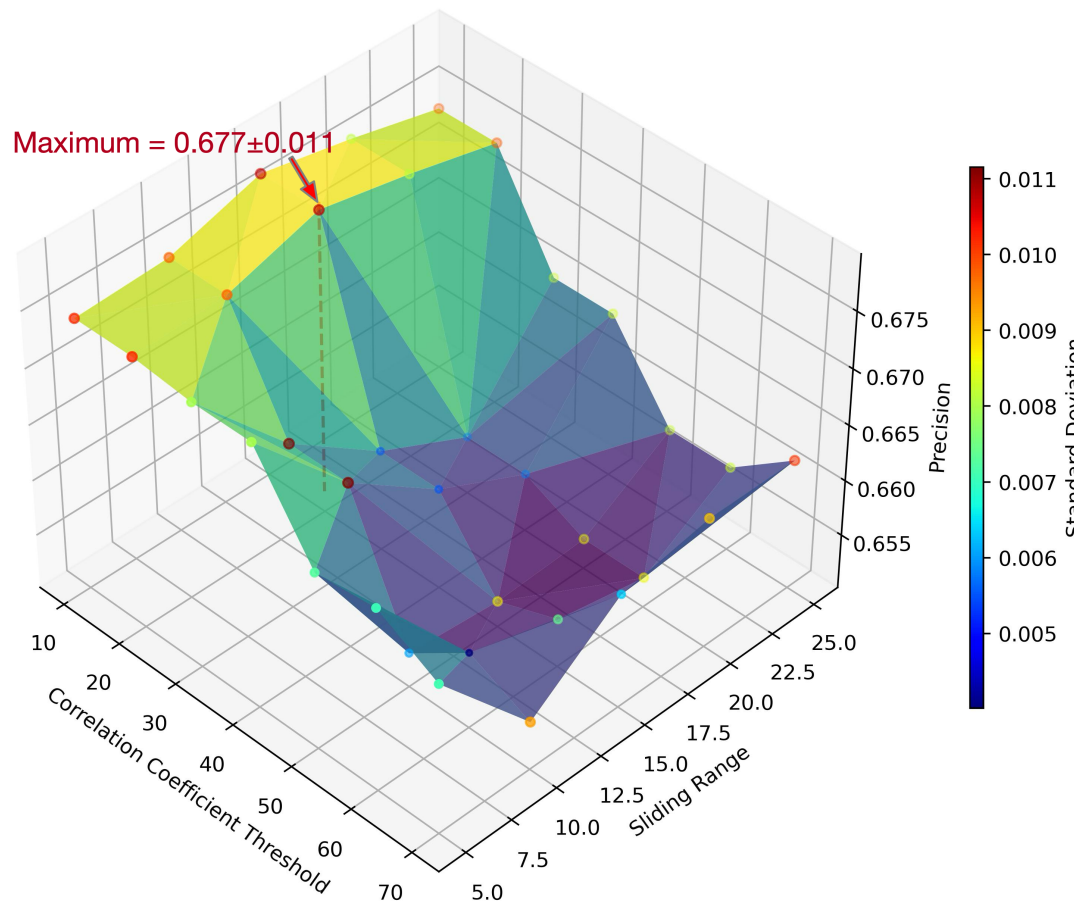

(b) Precision

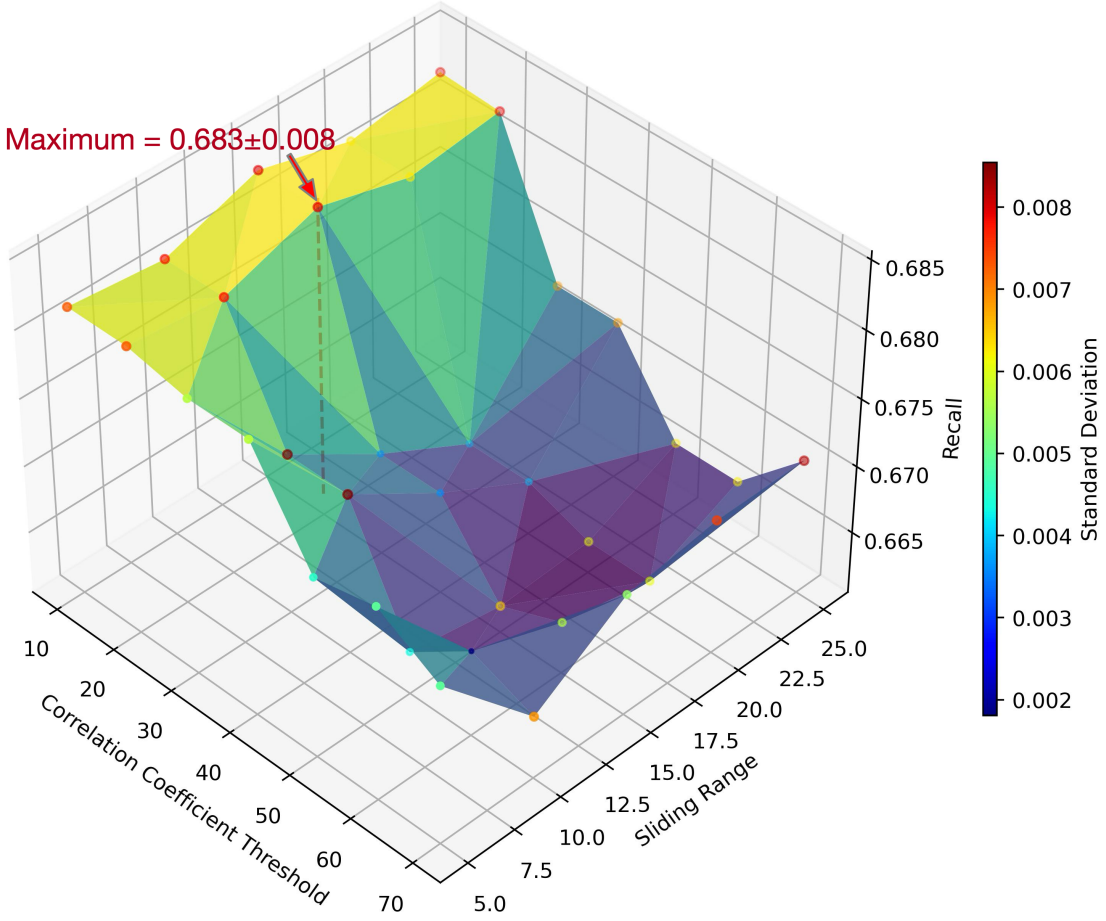

(c) Recall

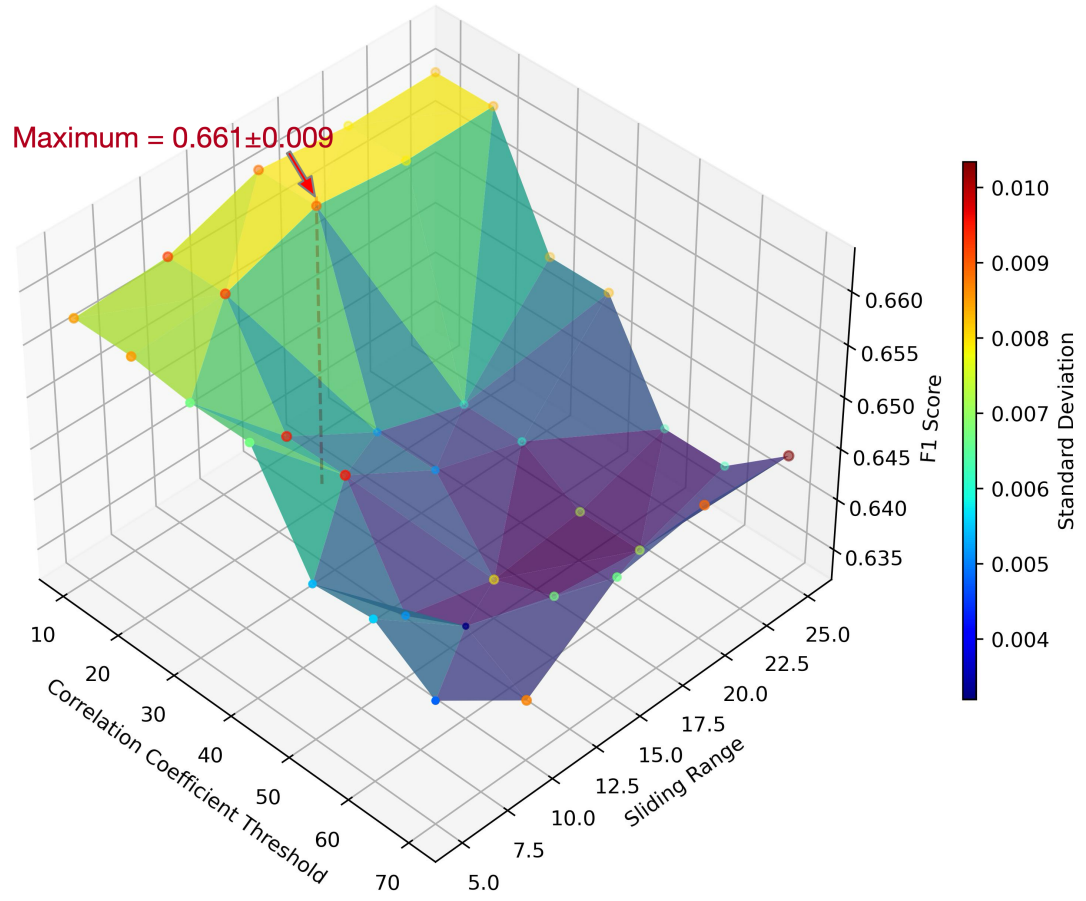

(d) F1 score

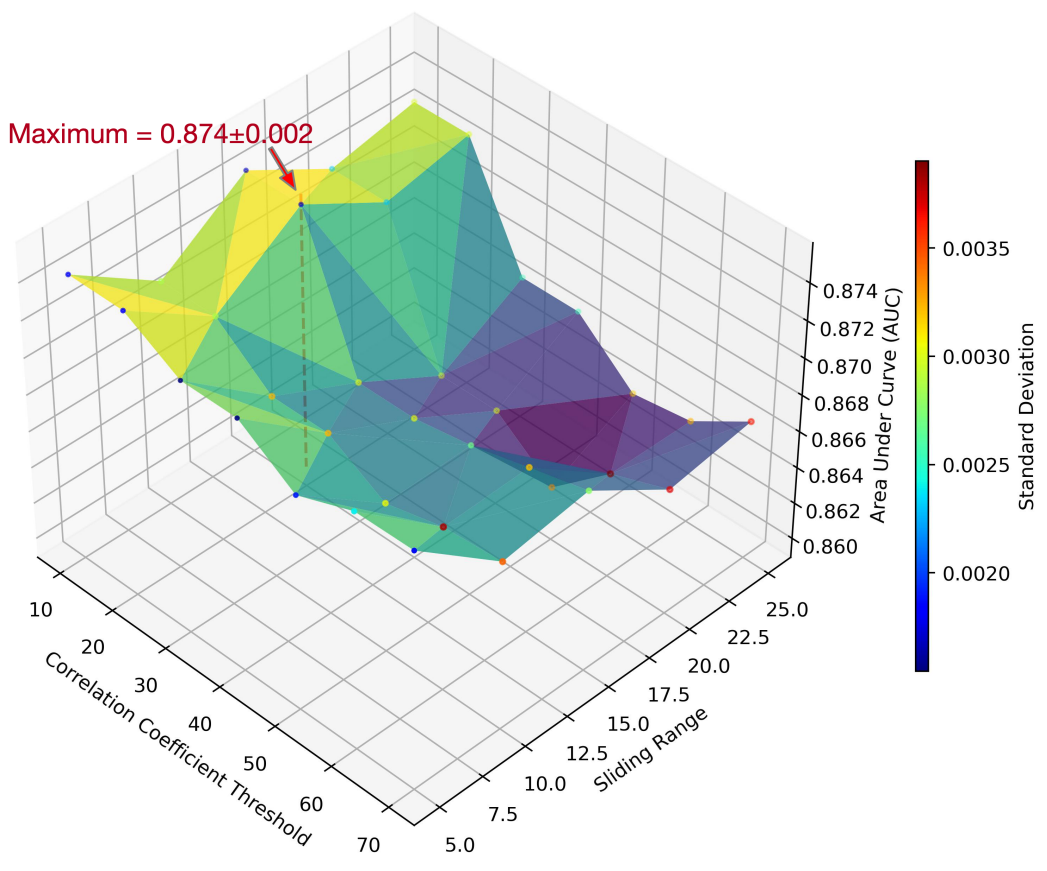

(e) Area Under Curve (AUC)

Figure 7 Filtering parameter searching according to prediction performance metrics (correlation coefficient threshold x0.01; maximum = mean ± standard deviation)

## 5.2 Analysis on key factors

Given the aforementioned parameter setting, a new rank of filtered factors can be generated. It is found that the Random Forest Classifier trained on the dataset with the first 15 factors on the rank attains the highest prediction performances which are represented by the maximum points in Figure 7. Therefore, these 15 factors are selected as the key factors. The histogram in Figure 8 presents the feature importance with respect to the key factors, and the descriptions of each factor are listed in Table 3. The feature importance is calculated using the method introduced in Section 4.2, which is indeed a global feature importance representing the contribution of a factor to all classes of the label. Besides, the feature importance of the key factors within each factor category is aggregated, and the pie charts in Figure 9 visualize the aggregated feature importance of the key factors in each factor category. Herein, Figure 9(a) presents the shares of factor categories with respect to the aggregated global feature importance, while Figures 9(b) to 9(d) respectively depict the shares of factor categories with respect to the aggregated feature importance for each class of the label. The primary analytical results regarding the aggregated global feature importance are summarized as follows:

1. The factors related to PSC deficiencies in the past years contribute most to the future incident likelihood. The category of PSC deficiencies is the top-ranked factor category in term of aggregated feature importance with three factors selected as key factors. This means that PSC deficiencies play a significant role in identifying incident risk of vessels. Addressing and rectifying PSC deficiencies is crucial for enhancing maritime safety and reducing the likelihood of incidents. Further, PSC deficiencies are direct indicators of the vessel's compliance with international maritime regulations and safety standards. The top-ranking of PSC deficiency factors suggests that adherence to these regulations significantly impacts the overall safety and incident likelihood of vessels.

2. Historical incidents and detentions are typically treated as two dominant categories that have effect on the likelihood of future vessel incident. However, as presented in Figure 9(a), those two categories seem not influential as the others in general in this case study. Also, the factors related to the performances of the management entities that the vessel belongs to, such as DOC company and flag state, seem more significant than those two categories. Specifically, comparing the factors F9 and F15 as presented in Figure 2, it can be found that both factors are decayed number of detentions in the past five years, while F9 represents the DOC company's performance and F15 reflects the individual vessel's performance, and the results demonstrate that F9 is more important than F15. This suggests that the incident likelihood might be more profoundly influenced by the regulatory compliance and management effectiveness of the entities, rather than the characteristics or experiences (e.g., incident and detention) of the vessel itself. Besides, the incidents or detentions occurring in the past recent years may impact the vessel's normal operation the subsequent year. That can to some degrees explains why the key factors related to incidents or detentions represent the performances in the past four or five year, instead of a more recent time period, such as two year like most other factors.

3. As presented in Table 3, it is found that most factors from the categories of historical performances are based on the cumulative performances over a time period in the past, rather than the annual performances. Although the factors of sailing (i.e., F3, F8, and F13) are annually measured, they represent the average sailing distance in the past first, second, and third years, respectively, which can also be seen as a cumulative factor with different weight assigned to each of the past three years. This finding demonstrates that the retrospective range of past years varies among different categories of factors. The impact of historical events (e.g., PSC deficiencies, incidents, etc.) on the future incident likelihood can indeed vary, and their durations of impact may differ. Some events may have a long-lasting impact, while the others may have a more immediate or short-term effect. Further, it is found that most historical performance factors adopt decay factors, which also demonstrates the difference in the aforementioned impact duration of events. Of note, there are two cumulative factors of PSC deficiencies without decay factors. It can be inferred that the three PSC deficiency factors can be integrated as an ensembled factor with a different combination of decay factors assigned to the past years, which suggests that optimizing the decay factors is a possible direction for future work.

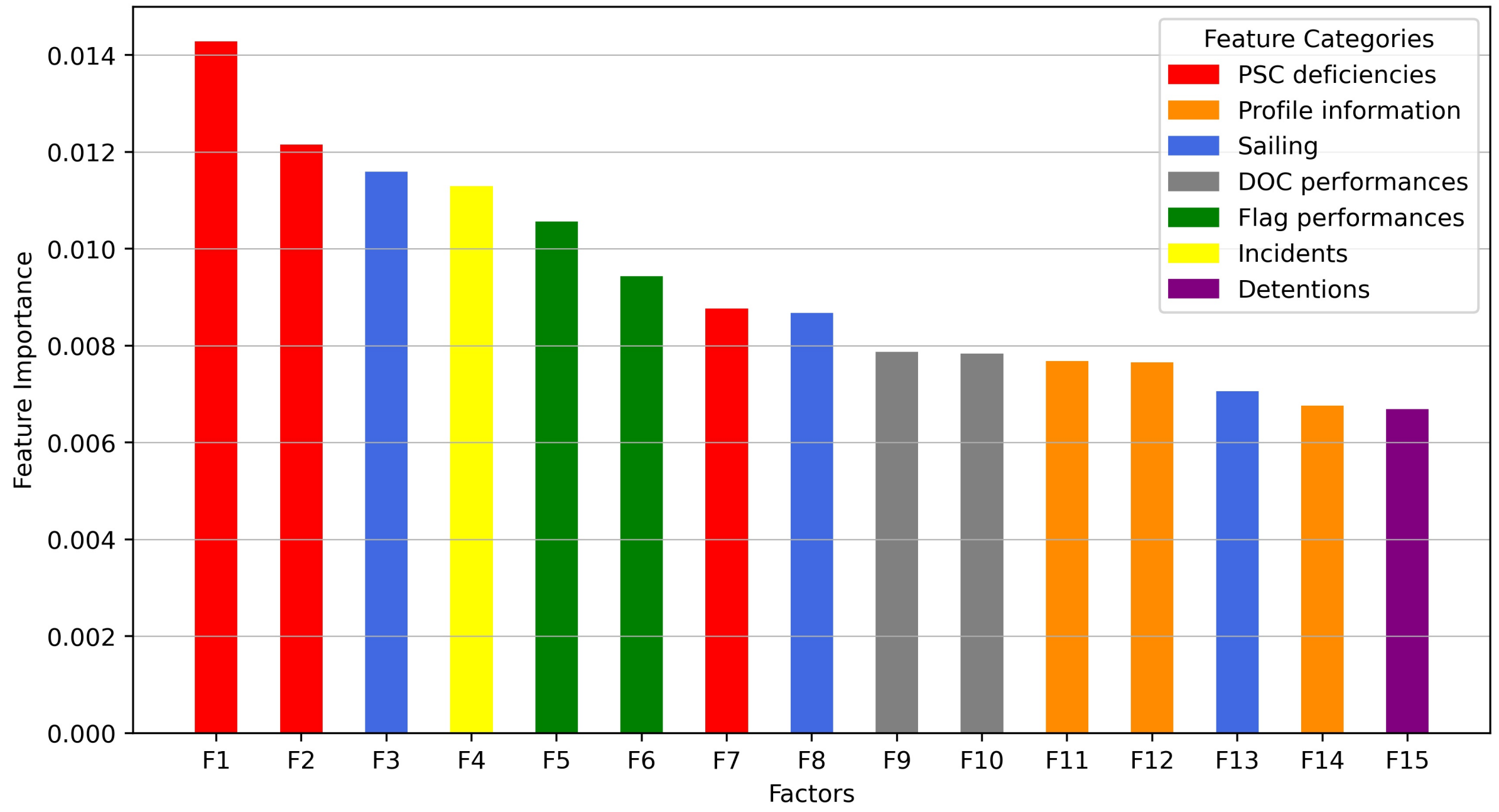

Figure 8 Feature importance with respect to key factors

Table 3 Descriptions of key factors

| Notations | Categories | Descriptions |
|---|---|---|
| F1 | PSC deficiencies | Decayed sum of PSC deficiencies in the past two years |
| F2 | PSC deficiencies | Number of deficiencies in the past four years |
| F3 | Sailing | Average sailing distance in the past year |
| F4 | Incidents | Decayed sum of severity across all the incidents in the past four years |
| F5 | Flag performances | Decayed number of red flags in the past five years |
| F6 | Flag performances | Decayed number of red flags in the past two years |
| F7 | PSC deficiencies | Number of deficiencies in the past two years |
| F8 | Sailing | Average sailing distance in the past second year |
| F9 | DOC performances | Decayed average detentions over the vessels that belong to the DOC company in the past five years |
| F10 | DOC performances | Decayed average deficiencies over the vessels that belong to the DOC company in the past five years |
| F11 | Profile information | Length Between Perpendiculars |
| F12 | Profile information | Daught |
| F13 | Sailing | Average sailing distance in the past third year |
| F14 | Profile information | Gross Tonnage |
| F15 | Detentions | Decayed number of detentions in the past five years |

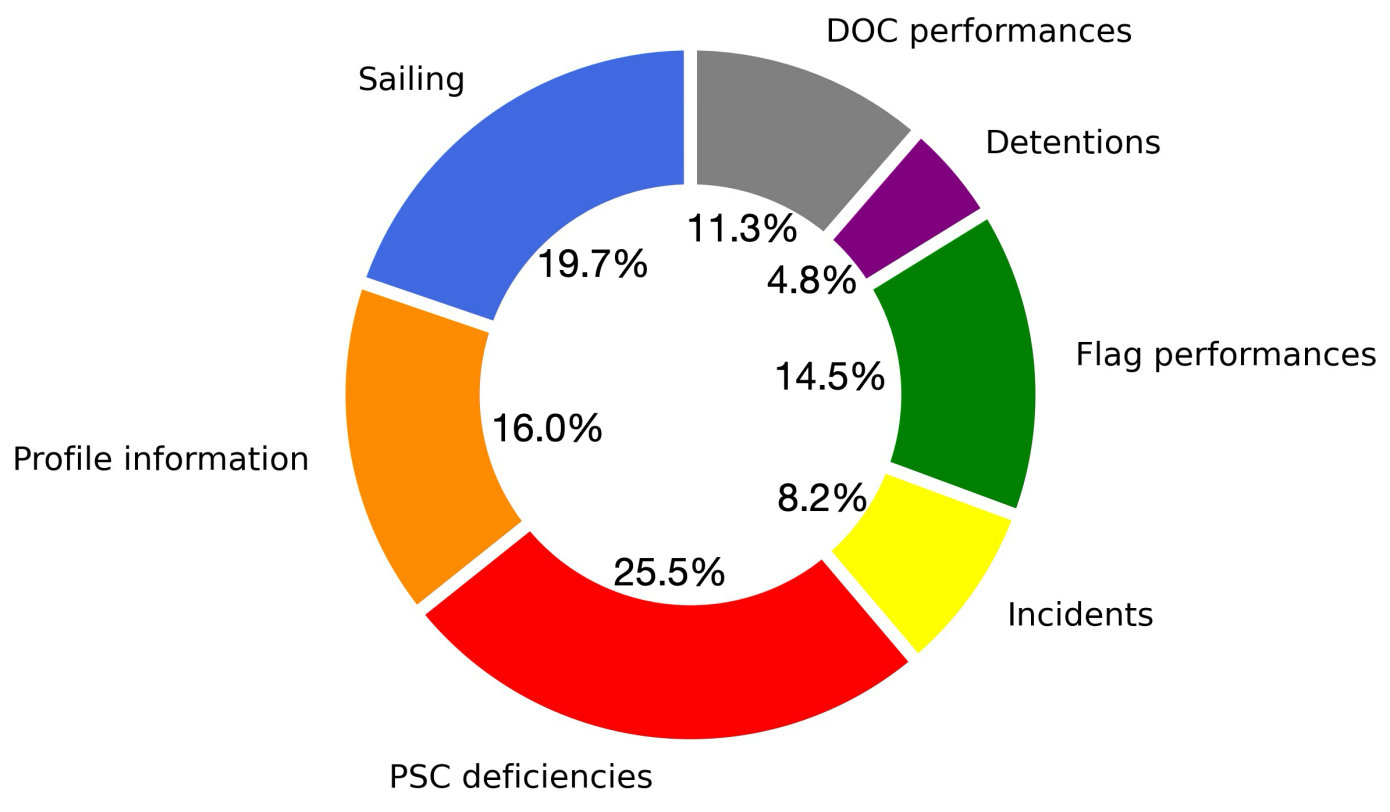

(a) Global (all classes)

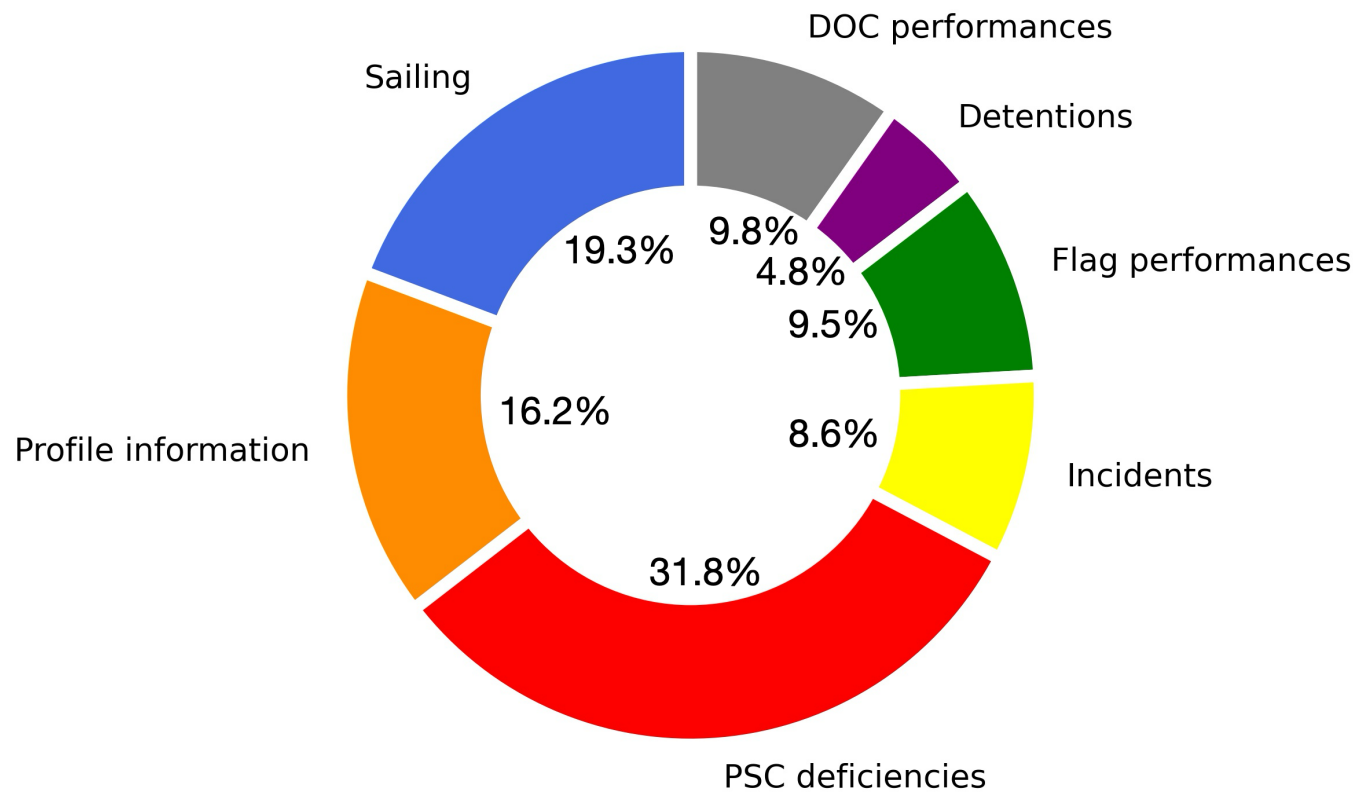

(b) Low incident risk level

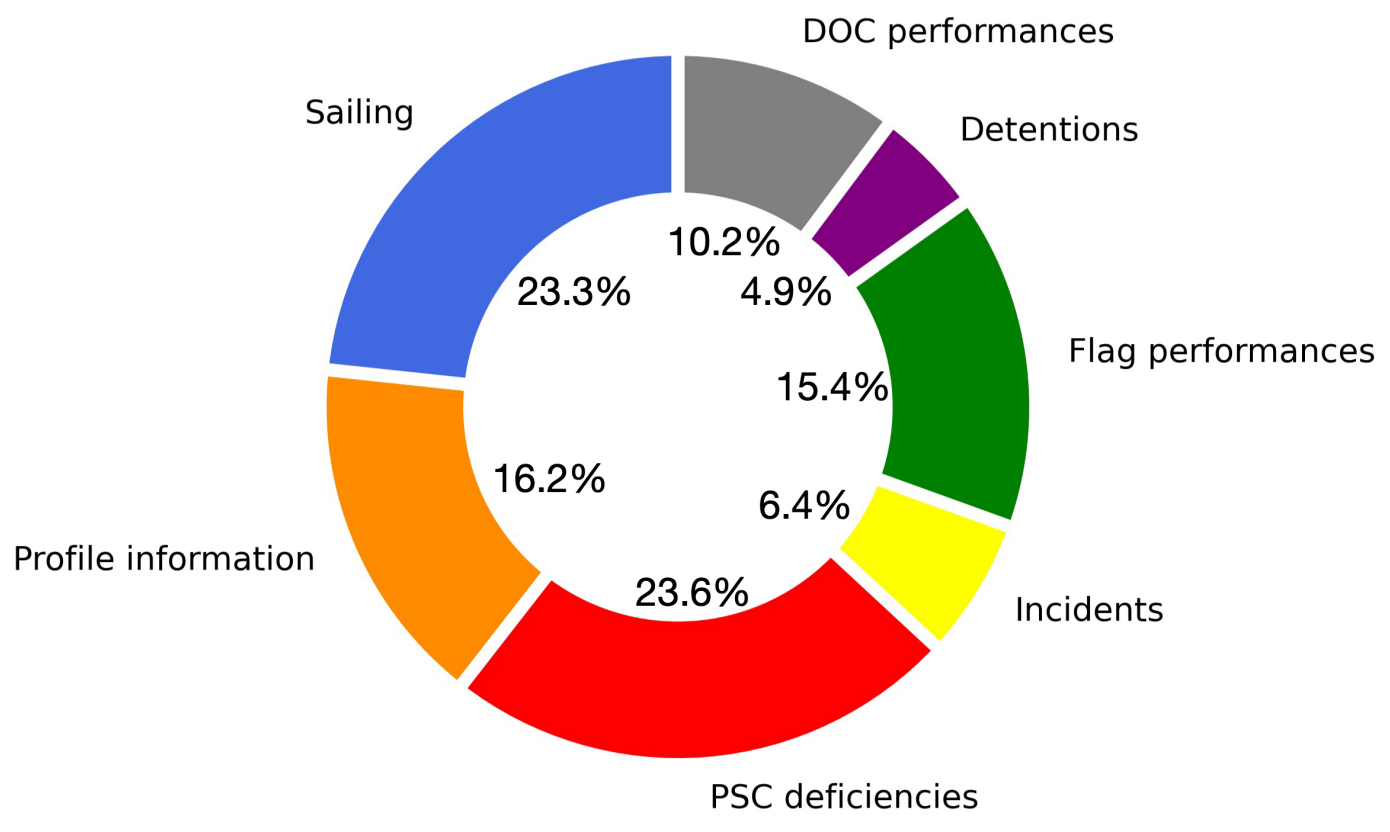

(c) Medium incident risk level

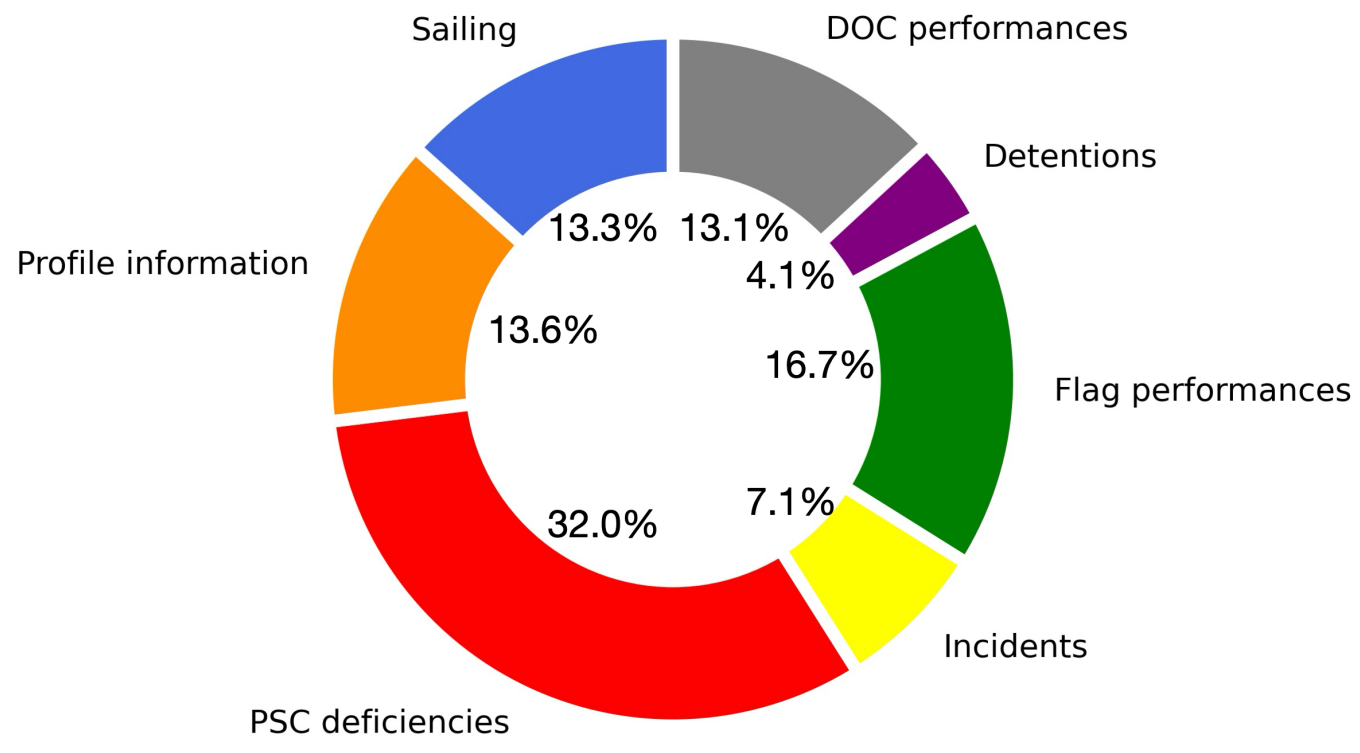

(d) High incident risk level

Figure 9 Aggregated feature importance with respect to factor categories

Figure 10 illustrates a detailed breakdown of the contribution of each key factor, as measured by SHAP value, to each class. In Figure 10, each data point represents a sample, with the color indicating the value of the factor for that particular sample, and the x-axis denotes the SHAP values of data points. The factors are ranked in descending order based on the average SHAP value of factor over the samples. Of note, a factor positively contributing to the low risk level implies its positive role in maintaining a safe vessel status, as the low risk level means that the vessel is involved in no incidents. The primary analytical results regarding the contributions of key factors to each class are summarized as follows:

1. As shown in Figures 9(b) and 9(d), the factors related to PSC deficiencies significantly contribute to the low and high incident risk levels. Specifically, as illustrated in Figures 10(a) and 10(c), the higher values of the PSC deficiency factors, which means a larger number of deficiencies detected in historical PSC inspection, negatively contribute to the low incident risk level and positively contribute to the high incident risk level. This suggests that PSC deficiencies could be an indicator for estimating incident likelihood, which is in line with the above finding related to the global feature importance of PSC deficiencies. Considering the contributions of PSC deficiency factors to incident risk, prioritizing efforts to address PSC deficiencies through improved inspection protocols, training, and enforcement could yield significant safety improvements.

2. As shown in Figures 9(b) to 9 (d), in addition to PSC deficiencies, the other two factor categories related to the performances of management entities (i.e., flag performances and DOC performances) also exhibit significant contributions to each incident level, as compared to the factor categories associated with the safety performances of individual vessel (i.e., incidents and detentions). Additionally, the percentage shares of both flag performances and DOC performances increase with the escalating incident risk levels. This further demonstrates the above clarified viewpoint that the regulatory compliance and management effectiveness of the entities may be more influential than individual vessel's experience on future incident likelihood.

3. Interstingly, as presented in Figures 10(a) to 10(c), it is found that a higher value of flag performance factors, which reflects inferior safety performance of flag state, is more likely to positively contribute to both low and high incident risk levels, while negatively contributing to the medium risk level. Upon reviewing the vessels with high-value flag performance factors, it is found that the values of their sailing-related factors are usually lower. This implies that the operation of these vessels may be restricted due to the poor safety performance of their vessel states according to certain protocols and regulations governing vessel states. Besides, Figures 10(a) and 10(b) indicate that sailing-related factors are more influential in determining both low and medium risk levels compared to flag performance factors, with lower sailing distance/time potentially decreasing the exposure of a vessel to risk. Thus, the lower sailing-related factors may counteract the effect of inferior flag performance to both low and medium risk levels. There are two-fold reasons that may explain why the high value of flag performance factor positively contribute to the high incident risk level. First, there are many vessels with inferior flag performances involved in severe incidents, even their operation time is limited. Second, as compared with the other two risk levels, the sailing-related factors are less dominant and sensitive to the high risk level, thereby making them less effective in counteracting the impact from the inferior flag performance.

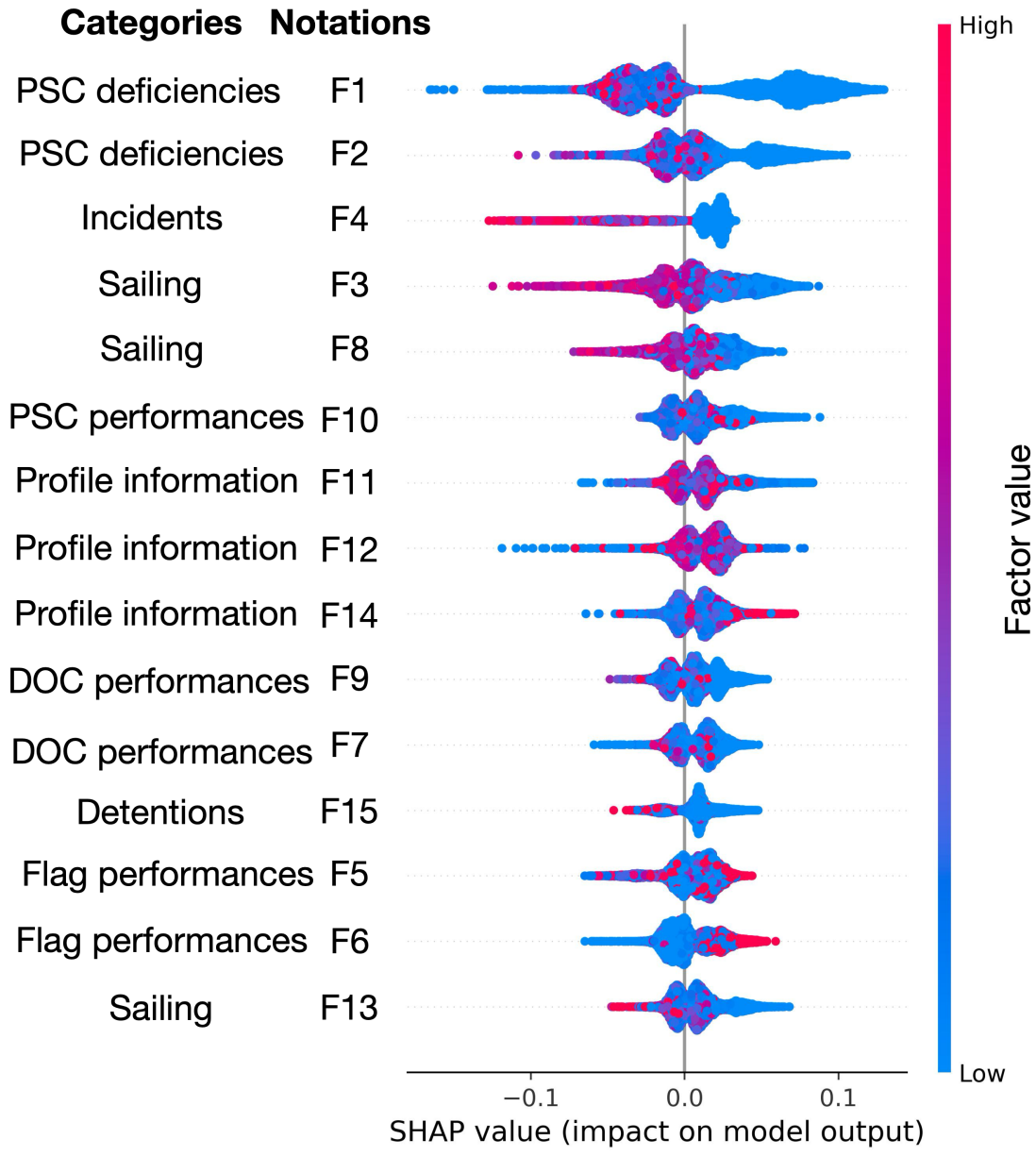

(a) Low incident risk level

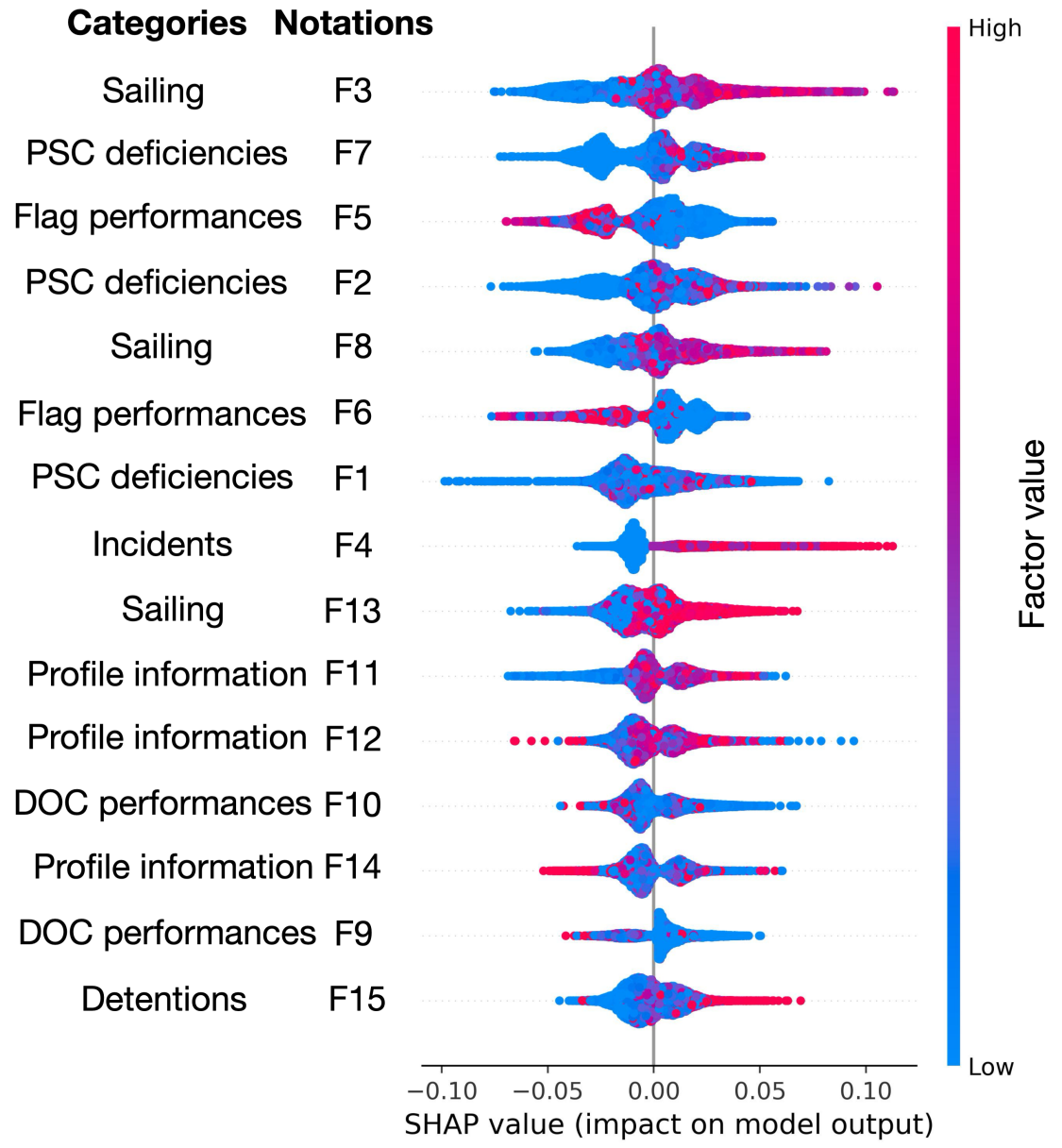

(b) Medium incident risk level

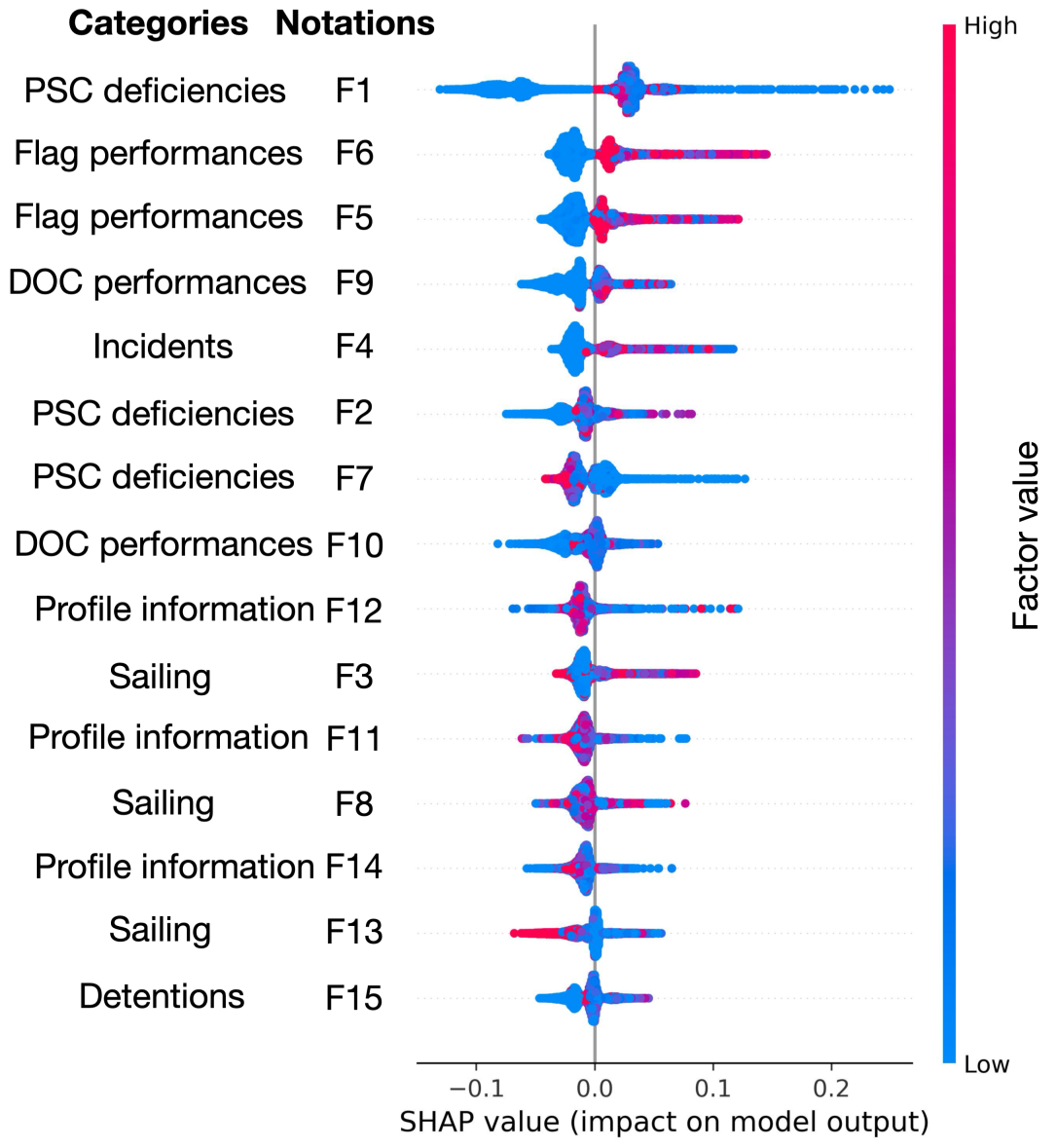

(c) High incident risk level

Figure 10 Feature importance with respect to label classes

### 5.3 Method comparison

In case study, the proposed feature selection method is compared to a conventional embedded method in terms of prediction performances and explainability of selected factors. As compared to the proposed method, the conventional embedded method also uses Random Forest classifier to be trained on the resampled data but without the feature filtering process. Figure 11 illustrates the prediction performances regarding the factor ranks generated by the two feature selection methods, respectively. Herein, the results shown in Figure 11 also explains how the maximum performances are obtained given the optimal combination of the two filtering parameters, as presented in Figure 7. As shown in Figure 11, as compared to the conventional feature selection method, the prediction performances achieved by the proposed feature selection method seem more satisfactory. Indeed, the relatively low prediction performances manifested in Figure 11 are expected, given that the maritime incidents are extremely-low-probability events, making them inherently challenging to predict accurately. Despite these challenges, the superiority of the proposed method over the conventional method in enhancing the prediction performances demonstrate that the factors selected by the proposed method are more rational and effective.

As shown in Figure 11(b), when the conventional embedded method is used to select factors, the first 14 factors on the rank are selected as key factors. Table 5 summarizes the categories and descriptions of the key factors selected by the conventional embedded method. It can be found that there are two main issues regarding the explainability in those selected factors. First, most of the selected factors exhibit high similarity to another; and second, the selected factors belong to only three categories, with a majority coming from the profile information category. Herein, as compared to the other factor categories, the relationship between profile information and the incident risk appears more as an association rather than a causal or direct contributory connection. Therefore, it is counterintuitive that the profile information factors dominate the incident risk. In summary, the above comparison results demonstrate that the proposed feature selection method is able to improve the performance of the incident risk prediction and ensure better explainability of the selected factors, as compared to the conventional embedded method.

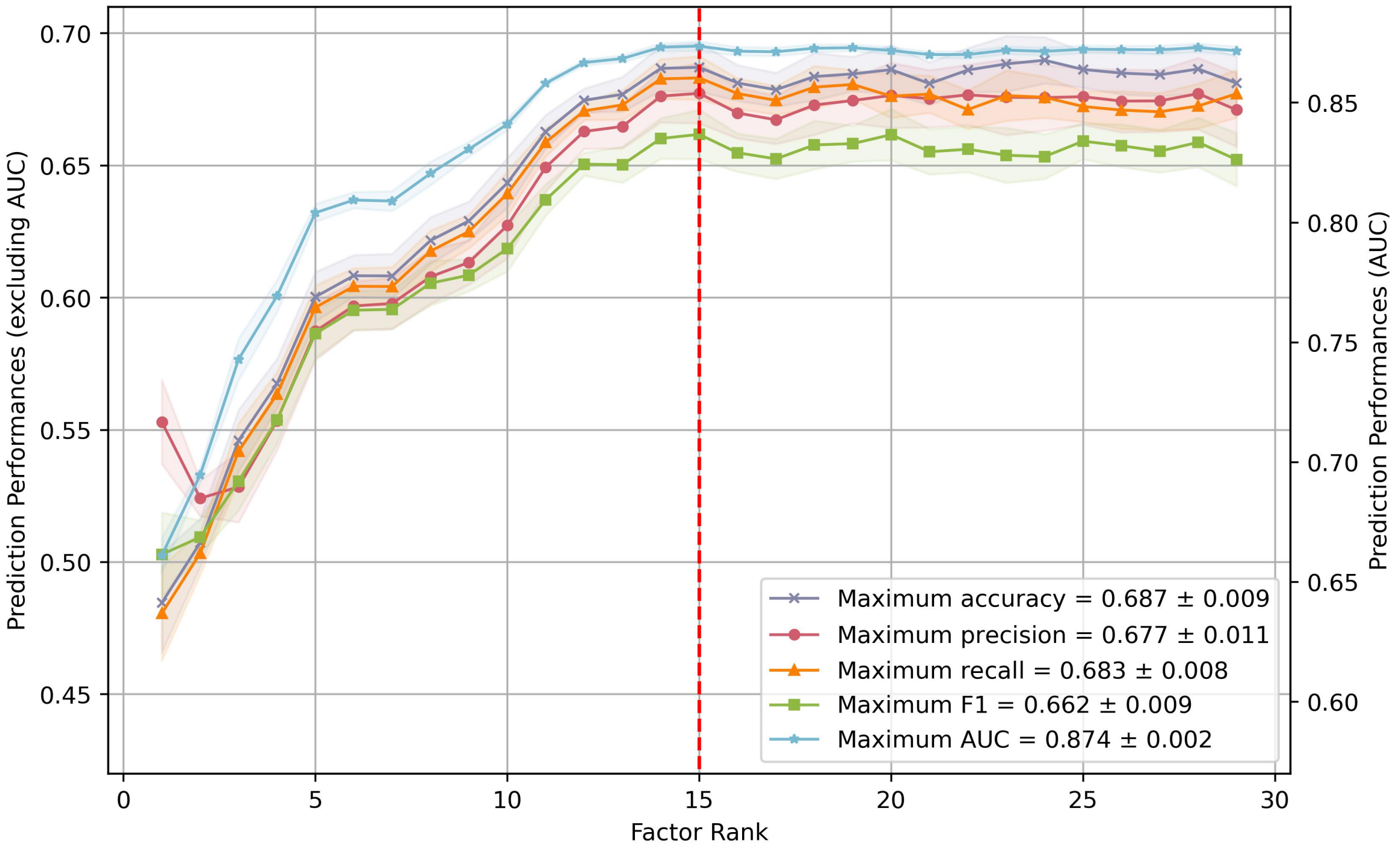

(a) Proposed method

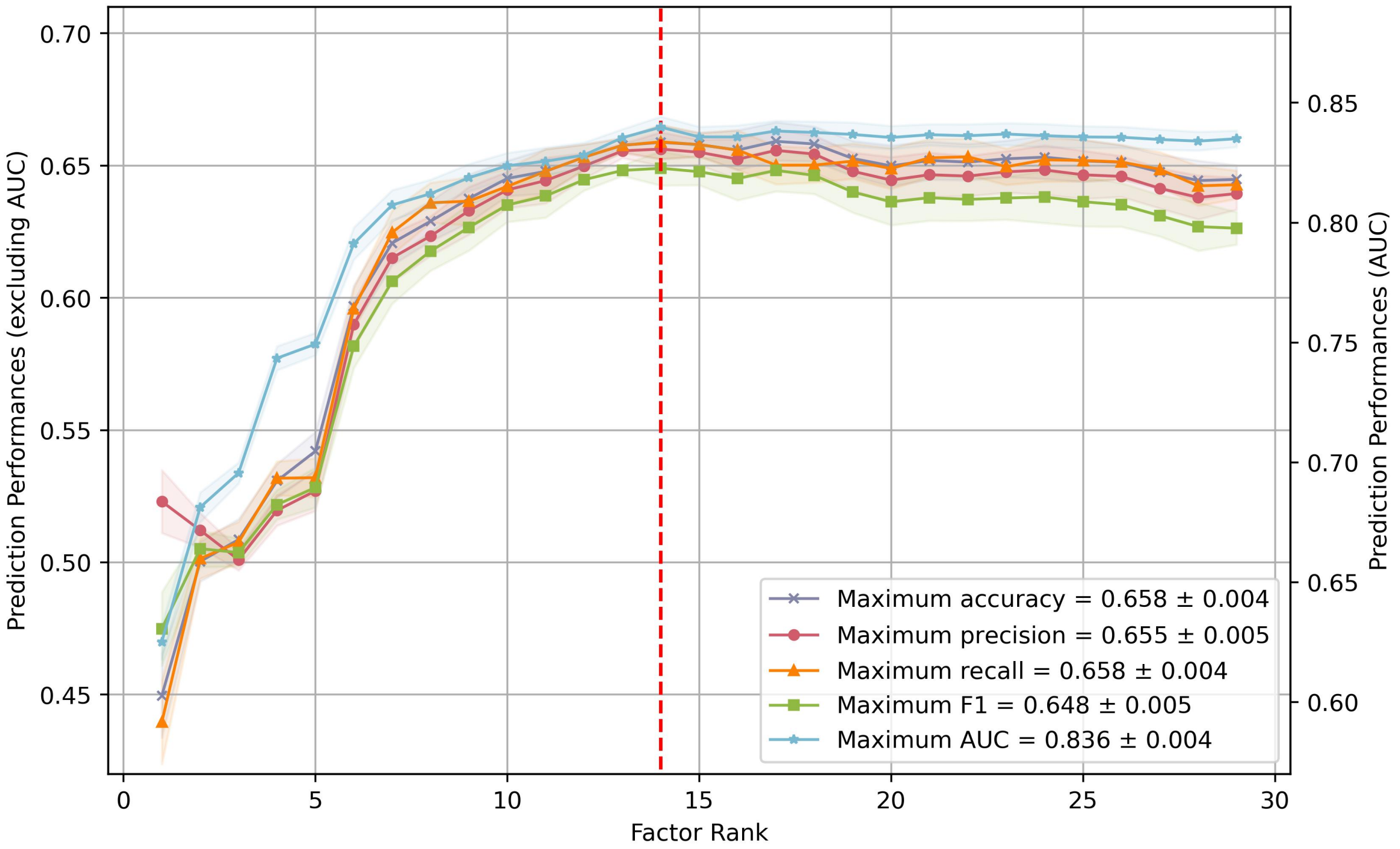

(b) Conventional embedded method

Figure 11 Cross-validation performances with respect to factor rank

Table 5 Rank of key factors selected by conventional embedded method

| Rank | Categories | Descriptions |
|---|---|---|
| 1 | PSC deficiencies | Decayed sum of PSC deficiencies in the past two years |
| 2 | PSC deficiencies | Decayed sum of PSC deficiencies in the past three years |
| 3 | PSC deficiencies | Decayed sum of PSC deficiencies in the past four years |
| 4 | Profile information | Daught |
| 5 | PSC deficiencies | Decayed sum of PSC deficiencies in the past four years |
| 6 | Profile information | Gross tonnage |
| 7 | Profile information | Deadweight tonnage |
| 8 | Profile information | Maximum dead weight tonnage |
| 9 | Profile information | Net tonnage |
| 10 | Profile information | Length between perpendiculars |
| 11 | Profile information | Depth |
| 12 | Profile information | Length overall |
| 13 | Sailing | Average sailing distance in the past year |
| 14 | Sailing | Average sailing distance over the past four year |

## 6. CONCLUSIONS

Identifying the key factors significantly contributing to the vessel incident risk in the long term is crucial for enhancing maritime safety. There are many factors that may potentially contribute to the incident risk, and each factor can be represented in multiple formats or styles. Besides, the contributions of the factors related to vessel's historical performances may decay at varying rate over time. Thus, feature selection becomes more essential, as it is able to reduce redundant and noisy factor, enhance factor interpretability, and improve model's prediction performance. This study aims to investigate the key factors that significantly contribute to the vessel incident risk in the subsequent year, based on a given datestamp. Herein, the risk label is defined as the risk levels graded by the total incident severity in the year. In this study, an improved embedded feature selection method with Random Forest classifier as the host machine learning model is proposed. Compared to the conventional embedded method, the proposed method incorporates a feature filtering process to reduce the highly-related factors. The vessel data utilized for the case study encompass the factors from seven categories, namely, incidents, PSC deficiencies, detentions, sailing, DOC performances, flag performances, and profile information. The results of the case study manifest the superior performances of the proposed feature selection method in incident risk prediction and factor explainability, as compared to the conventional embedded method. The satisfactory prediction performances achieved by the proposed method also reflect, to some degrees, the rationality and effectiveness of the selected factors. The primary findings from the selected key factors are summarized as follows: 1) Historical PSC deficiencies have the most significant contribution to the future incident risk, compared to the other categories of factors; 2) as compared to the risk experiences (i.e., historical incidents and detentions) of the vessel per se, the regulatory compliance and management effectiveness of the management entities (i.e., DOC company and flag state) that the vessel belongs to exert a more substantial impact on the future incident risk; and 3) the vessel's historical events (e.g., PSC deficiencies, detentions, and incidents) may exhibit different post-event impacts on the future incident risk, including the variations in impact duration and impact strength, and thereby their contributions to the incident risk may decay differently over time.

Those findings can have significant implications for the maritime industry. For example, based on the findings, maritime stakeholders can develop and implement enhanced risk management strategies by prioritizing the correction of historical PSC deficiencies. Besides, vessel operators and management entities investing in effective regulatory oversight and management practices can contribute significantly to incident prevention. In the future, there are several directions to resolve the limitations of this study and further improve the proposed feature selection method. First, maritime incidents are extremely-low-probability events, and thus, achieving satisfactory incident risk prediction results may always be a challenge. However, efforts can still be directed towards maximizing the prediction performance, such as addressing class imbalance problem more effectively and improving machine learning models. Second, this study considers all the incident types when selecting key factors, and in the future, feature selection method could be designed dedicated to a particular incident type, offering deeper insights into each incident type through factor analysis. Last but not least, more factors could be incorporated, and the factors could be better constructed for feature selection in the future work. For example, for the factors related to historical events, the setting of the decay factors could be optimized. Moreover, the underlying interactions between factors, which may not be easily revealed by correlations, such as the relationship between flag performances and sailing as aforementioned, could be further investigated during feature engineering.

## ACKNOWLEDGMENTS

This study is supported by the Singapore Maritime Institute via the project SMI-2022-MTP-09. Any opinions, findings and conclusions or recommendations expressed in this study are those of the author(s) and do not reflect the views of the Singapore Maritime Institute and RightShip. The authors do not have permission to share data.

## APPENDIX A SUMMARY OF PREVIOUS LITERATURE

Table A1 Categorization of previous maritime incident studies

| References | Factor categories | Label definitions | Analytical methodologies | Scopes[1] |
|---|---|---|---|---|
| [5] | Vessel particulars and previous histories[2] (incidents and inspections) [3] | Incident severity level | Hazard rate model | |
| [24] | Vessel particulars, environmental factors[4], and navigational factors | Incident severity level | Bayesian network | In coastal sea areas |
| [25] | Incident particulars, vessel particulars, human factors, and environmental factors | Incident severity level | Ordered logistic regression model | |
| [26] | Human factors and environmental factors | Incident severity level | Random Forest | Collision incidents |
| [27] | Human factors, vessel particulars, environmental factors, management, and port facilities | Incident severity level | Ordered logistic regression model | Berthing incidents |
| [28] | Vessel particulars, environmental factors, human factors, and management | Incident severity level | Bayesian Network | |
| [29] | Vessel particulars, incident particulars, and environmental factors. | Incident severity level | Bayesian network | |
| [30] | Vessel particulars, previous histories, flag, classification society, and DOC company. | Incident severity (cost) | Hazard rate model | |
| [32] | Vessel particulars, environmental factors, human factors, and incident particulars | Incident severity (injury) | Zero-inflated ordered probit model | |
| [33] | Environmental factors, Management, and vessel particulars | Incident types | Human Factor Analysis and Classification System | Collision incidents |
| [34] | Human factors | Incident types | Random Forest | |
| [35] | Vessel particulars, environmental factors, human factors, and management | Incident types | Bayesian Network | |
| [36] | Human factors (operations) and environmental factors | Incident types | Bayesian Network and Association Rule Mining | Fishing vessels |
| [37] | Vessel particulars, environmental factors, navigational factors, and human factors | Incident types | Bayesian Network | |
| [16] | Incident particulars, vessel particulars, and environmental factors (weather) | Incident types | Random Forest and Naïve Bayes | In Arctic |
| [38] | Incident particulars | Incident types (determined by clustering) | Kernel Density Estimation and K-means clustering | |
| [39] | Human factors, vessel particulars, environmental factors, and management | Incident types (determined by clustering) | Latent Dirichlet Allocation | |
| [40] | Incident particulars, vessel particulars, and environmental factors (weather). | Incident types and incident consequences (e.g., fatalities, injuries, etc.) | Structural Equation Modeling | |
| [41] | Incident particulars, environmental factors (weather), and geographical information | Incident types and incident consequences | Bayesian Network | In coastal sea areas |
| [42] | Environmental factors and human factor | Incident types and incident consequences | Bayesian Network | In Arctic |
| [43] | Incident particulars and vessel particulars | Incident consequences (injuries) | Poisson regression | |
| [44] | Navigational factors | Incident-prone sea area or not | Co-occurrence analysis | Collision incidents |
| [45] | Geographical information | Incident-prone sea area or not and area's incident severity | Multiple machine learning models | In coastal sea areas |
| [46] | Vessel particulars, navigational factors, and environmental factors (weather). | Incident occurrence | Support Vector Machine | In extreme weather events |
| [47] | Navigational factors, environmental factors, and geographical information | Incident occurrence | Long-Short-Term Memory neural network | |
| [7] | Previous histories, flag, DOC company, and vessel particulars | Incident risk level | Bayesian Network | |
| [48] | Human factors, vessel particulars, environmental factors, and management | Incident risk level | Failure Modes and Effects Analysis and Bayesian Network | Risks in human evacuation from cruise ships |
| [30] | Vessel particulars, previous histories, and environmental factors (oceanographic factors) | Binary classes based on incident probability | Binary regression model | |
| [8] | Vessel particulars, previous histories, and shipyard countries | Binary classes based on incident probability | Balanced Random Forest | |
| [49] | Human factors and environmental factors | Risk probability | Success Likelihood Index Method | Maritime operational risk (bunker spills) |
| [50] | Vessel particulars, flag, classification society, and geographical information | Risk probability | Combination of Bayesian rule and the least-squares method | Collision incidents |
| [51] | Workflow, human factors (operations), and safety barriers | Risk probability | Probabilistic network | Maritime operational risks |

1 1. The scopes includes the specific incident types, vessel types, regions/areas, etc. that are investigated in the paper.

2 2. The environmental factors in this table include weather, sea condition, visibility, etc. when an incident occurred.

3 3. In this table, the category followed a bracket means the paper only considers the factors listed within the bracket of this category.

4 4. The previous histories in this table denotes the records of historical incidents, detentions, and inspection deficiencies of a vessel.